 \documentclass[final,1p,times,twocolumn]{elsarticle}
\usepackage{amsmath,amssymb}
\usepackage{booktabs}
\usepackage{multirow}
\usepackage{longtable}
\usepackage{siunitx}
\usepackage{hyperref}
\journal{arXiv}
\biboptions{sort&compress}
\begin{document}

%%\begin{table*}[!t]
%%\ifpreprint\else\vspace*{-15pc}\fi
%%
%%\section*{Research Highlights}
%%
%%
%%\vskip1pc
%%
%%\fboxsep=6pt
%%\fbox{
%%\begin{minipage}{.95\textwidth}
%%
%%\vskip1pc
%%\begin{itemize}
%%
%% \item Studying main approaches of temporal modeling in deep-based human action recognition.
%% \item Methods are grouped into motion-based, RNNs, 3D filters, transformers, and hybrid.
%% \item In each category, methods are grouped based on used visual data modalities.
%% \item A comprehensive survey of transformer-based human action recognition is provided.
%% \item Some suggestions are proposed for future research using transformers.
%%
%%\end{itemize}
%%\vskip1pc
%%\end{minipage}
%%}
%%
%%\end{table*}
%%
%%\clearpage

\begin{frontmatter}

%% Title, authors and addresses

%% use the tnoteref command within \title for footnotes;
%% use the tnotetext command for theassociated footnote;
%% use the fnref command within \author or \address for footnotes;
%% use the fntext command for theassociated footnote;
%% use the corref command within \author for corresponding author footnotes;
%% use the cortext command for theassociated footnote;
%% use the ead command for the email address,
%% and the form \ead[url] for the home page:
%% \title{Title\tnoteref{label1}}
%% \tnotetext[label1]{}
%% \author{Name\corref{cor1}\fnref{label2}}
%% \ead{email address}
%% \ead[url]{home page}
%% \fntext[label2]{}
%% \cortext[cor1]{}
%% \affiliation{organization={},
%%             addressline={},
%%             city={},
%%             postcode={},
%%             state={},
%%             country={}}
%% \fntext[label3]{}

\title{Transformers in Action Recognition: A Review on Temporal Modeling}

%% use optional labels to link authors explicitly to addresses:
%% \author[label1,label2]{}
%% \affiliation[label1]{organization={},
%%             addressline={},
%%             city={},
%%             postcode={},
%%             state={},
%%             country={}}
%%
%% \affiliation[label2]{organization={},
%%             addressline={},
%%             city={},
%%             postcode={},
%%             state={},
%%             country={}}

\author[a,b]{Elham Shabaninia \footnote{Corresponding author. Email: e.shabaninia@kgut.ac.ir}}%
\affiliation[a]{organization={Department of Applied Mathematics, Faculty of Sciences and Modern Technologies, Graduate University of Advanced Technology},%Department and Organization
            %addressline={}, 
            city={Kerman},
            postcode={7631818356}, 
            %state={},
            country={Iran,}, 
            %email={  email: e.shabaninia@kgut.ac.ir}
	}%
\author[b]{Hossein Nezamabadi-pour \footnote{Email: nezam@uk.ac.ir}}%
\affiliation[b]{organization={Department of Electrical Engineering, Shahid Bahonar University of Kerman},%Department and Organization
            %addressline={}, 
            city={Kerman},
            postcode={76169133}, 
            %state={},
            country={Iran,},
           % email={  email: nezam@uk.ac.ir}
	}%
\author[c]{Fatemeh Shafizadegan \footnote{Email:  fatemeh.shafizadegan.1990@eng.ui.ac.ir}}%
\affiliation[c]{organization={Department of Computer Engineering, University of Isfahan},%Department and Organization
            %addressline={}, 
            city={Isfahan},
            postcode={8174673441}, 
            %state={},
            country={Iran,}, 
            %email={  email: fatemeh.shafizadegan.1990@eng.ui.ac.ir}
	}%

\begin{abstract}
%% Text of abstract
In vision-based action recognition, spatio-temporal features from different modalities are used for recognizing activities. Temporal modeling is a long challenge of action recognition. However, there are limited methods such as pre-computed motion features, three-dimensional (3D) filters, and recurrent neural networks (RNN) for modeling motion information in deep-based approaches. Recently, transformers' success in modeling long-range dependencies in natural language processing (NLP) tasks has gotten great attention from other domains; including speech, image, and video, to rely entirely on self-attention without using sequence-aligned RNNs or convolutions. Although the application of transformers to action recognition is relatively new, the amount of research proposed on this topic within the last few years is astounding. This paper especially reviews recent progress in deep learning methods for modeling temporal variations. It focuses on action recognition methods that use transformers for temporal modeling, discussing their main features, used modalities, and identifying opportunities and challenges for future research. 
\end{abstract}

%%Graphical abstract
%\begin{graphicalabstract}
%%\includegraphics{grabs}
%\end{graphicalabstract}

%%Research highlights
%\begin{highlights}
%\item Research highlight 1
%\item Research highlight 2
%\end{highlights}

\begin{keyword}
transformer \sep action recognition \sep deep learning \sep temporal modeling 
%% keywords here, in the form: keyword \sep keyword

%% PACS codes here, in the form: \PACS code \sep code

%% MSC codes here, in the form: \MSC code \sep code
%% or \MSC[2008] code \sep code (2000 is the default)

\end{keyword}

\end{frontmatter}

%% \linenumbers

%% main text
\section{Introduction}\label{Intro}
Video-based action recognition is the task of recognizing human activities (including gestures, simple actions, human-object/human-human interactions, group activities, behaviors, and events) from video sequences or still images \cite{RN1,RN2}. Compared with video-based methods, human activity recognition (HAR) from static images is still an open and challenging task \cite{RN3,RN4,RN5} and includes a limited range of proposed methods. Due to many applications, vision-based human action recognition is known as an old field of computer vision and different data modalities are adopted for recognition in the literature, including RGB, depth, skeleton, infrared, point cloud, etc. while the three first modalities are used primarily for human action recognition. RGB data provides the details of a scene (including shape, color, and texture) and helps describe the semantics of actions, depth maps provide three-dimensional (3D) structural information about the scene. On the other hand, skeletal data is high-level information about the 3D location of joints. Multi-modal approaches use the knowledge of different modalities for the visual understanding of complex actions.

Today, human action recognition methods are mainly established with the help of deep neural networks (DNNs) \cite{RN6,RN7,RN8,RN9,RN10,RN11,RN12,RN13,RN14,RN15,RN16}. That is mainly due to the success of convolutional neural networks (CNNs) in encoding spatial information of images for object detection and recognition. Various studies discovered the abilities of CNNs in automatically extracting useful and discriminative features from images that generalize very well \cite{RN17,RN18,RN19,RN20}. In addition, deep networks can scale up to tens of millions of parameters and huge labeled datasets \cite{RN8}.  Consequently, the computer vision community mainly focused on using the capacity of deep architectures in almost all fields of research, including human action recognition. However, besides encoding spatial information of frames, video analysis is involved with modeling temporal information. 

Encoding temporal information is of vital importance in recognizing different subtle sub-activities. Each activity is divided into different sub-activities. The sequence of these sub-activities differentiates among different activities. However, the temporal dimension typically causes action recognition to be challenging. On the other hand, existing deep architectures generally encode temporal information with limited solutions \cite{RN21,RN22,RN23,RN24} such as 3D filters, pre-computed motion features, and recurrent neural networks (RNNs). These models are typically restricted in simultaneously acquiring local and global variations of temporal features. 

On the other hand, the transformer is a new encoder-decoder architecture that uses the attention mechanism to differentially weigh each part of the input data \cite{RN25}. Although transformers are designed to handle sequential input data, they do not necessarily process it in order. Rather, the attention mechanism provides context for any position in the input sequence. This feature allows for more parallelization than RNNs and therefore reduces training times. Transformers achieved great success in natural language processing (NLP) tasks \cite{RN25} and are now applied to images \cite{RN26,RN27,RN28}.
 
\begin{figure}
\centering
\includegraphics[width=0.98\columnwidth]{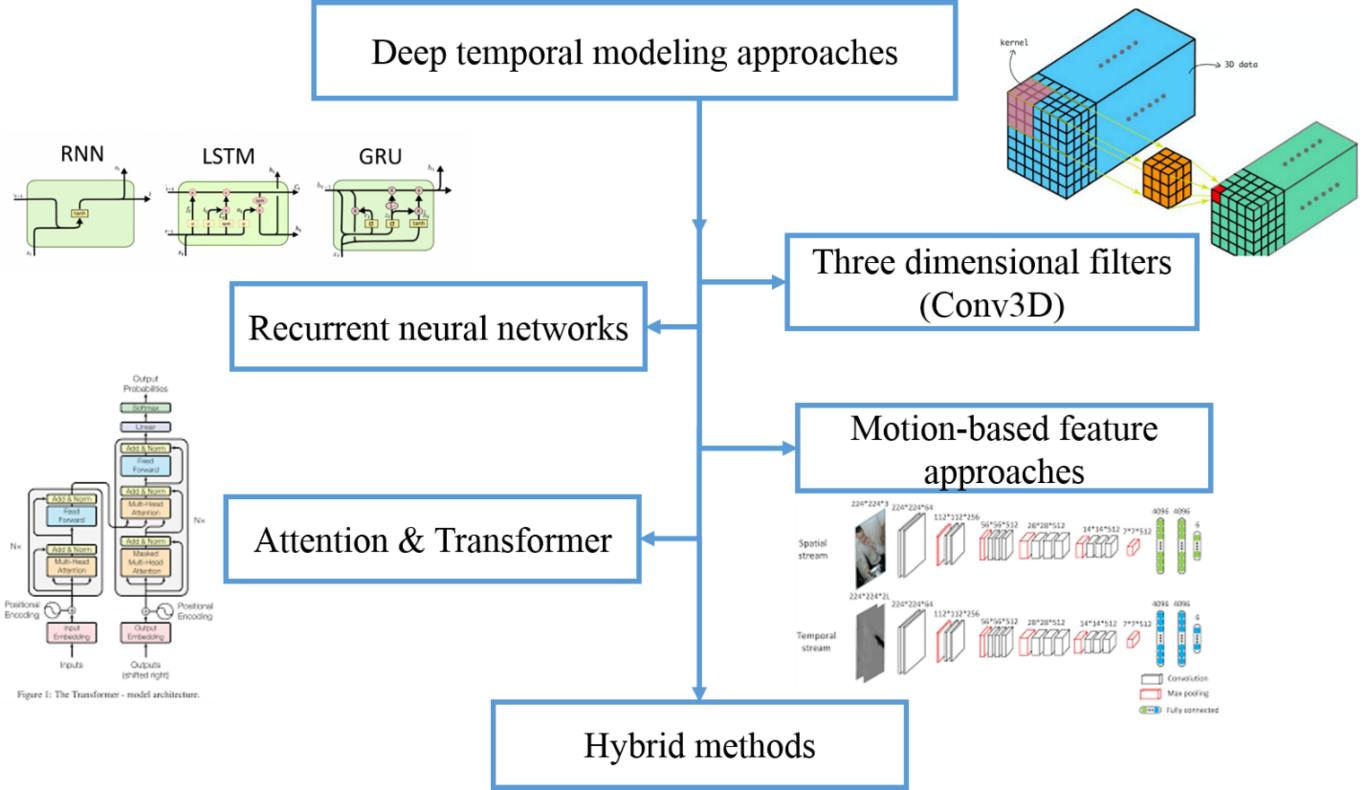}
\caption{The taxonomy of this paper: different methods are categorized into five approaches.}
\label{fig:image1}
\end{figure}

Along with NLP, video recognition is a perfect candidate for transformers, where videos are represented as a sequence of images (similar to language processing, in which the input words or characters are represented as a sequence of tokens \cite{RN29}). The application of transformers for action recognition is relatively new. However, the amount of research proposed on this topic within the last few years is increasing. This paper aims at capturing a snapshot of trends for temporal modeling in human action recognition. It focuses on supervised learning methods that often require a large amount of data with expensive labels for training models. Meanwhile, unsupervised and semi-supervised learning techniques \cite{RN30,RN31} (which enable to leverage of the availability of unlabeled data to train the models) are beyond the scope of this paper. The methods are categorized into five groups: motion-based feature approaches, three-dimensional convolutional neural networks, recurrent neural networks, transformers, and hybrid methods (see Figure \ref{fig:image1}). This categorization is advised by some research approaches. Especially, three first categories of pre-computed motion features, 3D filters, and RNN are mentioned in \cite{RN21,RN22,RN23,RN24}. However, transformers for video action recognition are a new approach. Finally, the hybrid category is designed to encompass combined methods.
So main contributions of this paper are as follows:
\begin{enumerate}
\item This paper reviews the main approaches proposed for modeling temporal information for human action recognition in deep-based methods. 
\item The deep learning-based approaches are categorized into motion-based, RNNs, 3D filters, transformers, and hybrid methods.  
\item In each category, methods are grouped based on used visual data modalities (RGB, depth, skeleton) to better compare similar approaches.
\item We provide a comprehensive survey of transformer-based human action recognition methods.%
\item Some suggestions are proposed for future research on human action recognition using transformers.
\end{enumerate}

The remainder of this paper is organized as follows. Section \ref{Related Survey Papers} reviews related survey papers on human action recognition. Section \ref{Temporal Modeling in Action Recognition Methods} provides a brief review of traditional approaches for temporal modeling and introduces five deep learning-based approaches for modeling the time dimension. These approaches are discussed in detail in different subsections. In each subsection, distinct modalities or combinations of multiple modalities used in methods are explored. Discussions and prospects are provided in section \ref{Discussions and Future Prospects}. The paper concludes in section \ref{Conclusions}.
%%%%%%%%%%%%%%%%%%%%%%%%%%%%%%%%%%%%%%%%%%%%%%%
\section{Related Survey Papers}
\label{Related Survey Papers}
Human action recognition is one of the old and interesting topics of computer vision. There are a lot of survey papers on this topic targeting different aspects of action recognition. Table \ref{tab:table1} lists some recent survey papers on human action recognition. As this table shows, some existing papers provide a review of both traditional and deep-based approaches, while others only concentrate on deep-based methodologies. On the other hand, there are some surveys on applications of human action recognition \cite{RN32,RN33} or benchmark datasets of action recognition \cite{RN15,RN34,RN35,RN36}. In addition, a group of reviews focuses on specific data modalities such as visual or sensor-based methods \cite{RN7,RN12,RN37,RN38,RN39,RN40,RN41,RN42}. While some others review approaches based on multiple data modalities \cite{RN9,RN43,RN44,RN45,RN46,RN47,RN48}. Compared with the existing survey papers:

\begin{enumerate}
\item This paper provides a novel taxonomy to review the main approaches of temporal modeling in human action recognition, with the main emphasis on transformer-based methods.
\item The survey includes both single-modal and multi-modal approaches of human action recognition.
\item Vision-based methods with RGB, depth, skeleton and hybrid features are considered.
\item A short review of conventional methods besides a long review of deep-based approaches is included.
\end{enumerate}

Note that some sections of this paper that consist of input modality + modeling approach are mentioned in some other survey papers. For example in \cite{RN49}, some approaches of skeleton + 3D filters, in \cite{RN27}, some approaches of RGB/Skeleton + transformer, and in \cite{RN15}, some approaches of RGB/depth + Motion features are reviewed. All these papers are comprehensive and informative. However, the interests of these papers are other topics (see Table \ref{tab:table1}) and there is no paper with the main emphasis on transformers and temporal modeling that reviews all these categories and modalities altogether. 
\begin{center}
\begin{table}
\scriptsize
\centering
\caption{Recent survey papers on human action recognition(C: Reviewing Conventional approaches, D: Reviewing deep-based approaches).}
\label{tab:table1}
\begin{tabular}{|c|c|p{9.5cm}|c|c|}

%\begin{tblr}{|Q[l,t]|Q[c,m]|Q[r,b]|}
\toprule
%References & Year & Main Focus & {\rotatebox[origin=c]{90}{Reviewing Conventional approaches }} & {\rotatebox[origin=c]{90}{Reviewing Deep-based approaches }}
References & Year & Main Focus & C & D
\\ \midrule\midrule
Yuanyuan et al. \cite{RN21} & 2021 & Categorizing deep approaches into Two-stream, 3D CNN, and LSTM groups. & \checkmark & \checkmark
\\ \midrule
Pareek et al. \cite{RN50} & 2021 & Reviewing features in action presentation, applications, and challenges in HAR. & \checkmark & \checkmark
\\ \midrule
Khan et al. \cite{RN22} & 2021 & Categorizing deep approaches into CNN, RNN, and hybrid. 
&\checkmark & \checkmark
\\ \midrule
Rangasamy et al. \cite{RN23} & 2020 & Categorizing deep-based architectures into CNN, 3D CNN, RNN, and LSTM (with a focus on sports video analysis). 
&\checkmark & \checkmark
\\ \midrule
Jegham et al. \cite{RN51} & 2020 & approaches are grouped into template-based, generative, and discriminative models.
&\checkmark & \checkmark
\\ \midrule
Zhang et al. \cite{RN14} & 2019 & Overviewing methods in action feature representation based on deep learning. 
&\checkmark & \checkmark
\\ \midrule
Dhiman et al. \cite{RN52} & 2019 & Investigating methods in abnormal human action recognition. 
&\checkmark & \checkmark
\\ \midrule
Estevam et al. \cite{RN53} & 2021 & Studying zero-shot video-based action recognition methods.  
&\checkmark & \checkmark
\\ \midrule
Zhu et al. \cite{RN24} & 2020 & Categorizing deep-based approaches into CNNs, two-stream, and 3D CNNs.
& & \checkmark
\\ \midrule
Yao et al. \cite{RN54} & 2019 & Studying CNN-based approaches.
& & \checkmark
\\ \midrule
Sreenu et al. \cite{RN55} & 2019 & Reviewing application of HAR in video surveillance for crowd analysis.
&\ & \checkmark
\\ \midrule
Chen et al. \cite{RN37} & 2021 & Surveying deep approaches in sensor-based methods.
& & \checkmark
\\ \midrule
Nguyen et al. \cite{RN38} & 2021 & Reviewing deep-based approaches and power requirements in mobile and wearable sensors.
& & \checkmark
\\ \midrule
Hussain et al. \cite{RN39} & 2020 & Categorizing methods into wearable, object-tagged, and device-free approaches. Further, device-free studies are grouped into action, motion, and interaction.
&\checkmark & \checkmark
\\ \midrule
Dang et al. \cite{RN7} & 2020 & Analyzing vision-based and sensor-based methods and their corresponding procedure of data collection, preprocessing, feature engineering, and training. 
&\checkmark & \checkmark
\\ \midrule
Beddiar et al. \cite{RN40} & 2020 & Reviewing vision-based approaches according to feature extraction process, recognition stage, source of input data, and machine learning supervision level.
&\checkmark & \checkmark
\\ \midrule
AI-Faris et al. \cite{RN41} & 2020 & Categorizing vision-based approaches using deep learning into generative and discriminative models.
&\checkmark & \checkmark
\\ \midrule
Wang et al. \cite{RN42} & 2019 & Analyzing ten approaches using conventional and deep strategies on visual modality (Kinect-based) 
&\checkmark & \checkmark
\\ \midrule
Wang et al. \cite{RN12} & 2018 & Classifying deep-based segmented and continuous motion recognition approaches into RGB, depth, skeleton, and hybrid.
& & \checkmark
\\ \midrule
Yadav et al. \cite{RN44} & 2021 & Deep and conventional approaches are grouped into vision-based, wearables, and multimodal categories.
&\checkmark & \checkmark
\\ \midrule
Majumder et al. \cite{RN43} & 2020 & Deep and conventional approaches are categorized into RGB, depth, and RGB \& depth.
&\checkmark & \checkmark
\\ \midrule
Rahmani et al. \cite{RN45} & 2021 & Deep approaches are categorized into visual modality and non-visual modality. 
& & \checkmark
\\ \midrule
Majumder et al. \cite{RN46} & 2020 & Traditional and deep approaches are grouped into fusions of RGB \& inertial, depth \& inertial, and RGB \& depth \& inertial. 
&\checkmark & \checkmark
\\ \midrule
Li et al. \cite{RN47} & 2020 & Methods in multi-user or group activity recognition are categorized into vision-based, sensor-based, radiofrequency, and hybrid groups.
&\checkmark & \checkmark
\\ \midrule
Liu et al. \cite{RN9} & 2019 & Deep and conventional methods with depth, skeleton, and hybrid features are considered. 
&\checkmark & \checkmark
\\ \midrule
Ulhaq et al. \cite{RN291} & 2022 &  The key contributions and trends to adapting Transformers
for visual recognition of human action are summarized. 
&\checkmark & \checkmark

\\ \bottomrule 	
\end{tabular}
\end{table}
\end{center}
%%%%%%%%%%%%%%%%%%%%%%%%%%%%%%%%%%%%%%%%%%%%%
\section{Temporal Modeling in Action Recognition Methods}
\label{Temporal Modeling in Action Recognition Methods}
Video-based action recognition is a video content analysis (VCA) task responsible for automatically analyzing the captured video to detect or recognize specific actions performed.  The critical issue in VCA is representing suitable spatio-temporal features and modeling dynamical patterns \cite{RN56}. VCA approaches are categorized into frame-by-frame-based methods and volumetric approaches. The former typically extracts a set of features from each frame. The features are then usually considered as time-series data. On the other hand, volumetric approaches implicitly model temporal dynamics and consider the video as a 3D volume. They extend standard features used for images to the 3D case.  

In recent decades, the vision community has suggested numerous action recognition techniques using RGB or depth. Among them, there are some promising methods, including representations for local spatio-temporal features \cite{RN57} such as SIFT3D \cite{RN58}, ESURF \cite{RN59}, HOG3D \cite{RN60}, HOF\cite{RN61}. These traditional action recognition methods use several detected salient points and local feature descriptors for each point. The local descriptors are then collected into a holistic descriptor for the entire video to be used for classification. The pro of using these local features is that they do not require detecting the human body, and the local features are almost robust to illumination changes, cluttered background, and noise. The con is the lack of semantics and limitation in discriminative capacity \cite{RN62}. Approaches such as Motionlets \cite{RN63}, Action Bank \cite{RN64}, Motion Atoms \cite{RN65}, Dynamic-Poselets \cite{RN62}, and Actons \cite{RN66} are proposed to account for these limitations. 

For modeling the temporal variation of skeletons, different approaches are proposed in the literature for traditional methods. In some approaches, the features computed from the action sequences are clustered into posture visual words (representing the prototypical poses of actions), and then the temporal evolutions of those visual words are modeled by explicit methods such as hidden Markov models (HMM) \cite{RN67,RN68} or conditional random fields (CRF) \cite{RN69,RN70}. Some other approaches consider the manifold of the trajectories \cite{RN71} or use hierarchical extended histogram (HEH) for modeling temporal variation of features acquired from individual frames of input sequence \cite{RN72}. These hand-crafted features and descriptors are recently substituted with deep representations to automatically extract high-level information from training data without using hand-crafted rules.

As mentioned above, in recent years there has been rapid development in deep learning-based methods for human action recognition.  Numerous studies are proposed in the literature for solving different challenges of human action recognition using deep architectures. Reviewing all these methods is a relatively comprehensive task. Many surveys discuss the pros and cons of different methods in detail \cite{RN6,RN7,RN8,RN12}. Here the focus is on how different methods deal with the temporal dimension. The methods are categorized into five groups: motion-based feature approaches, three-dimensional convolutional neural networks, recurrent neural networks, transformers, and hybrid methods. These methods are discussed in detail in the following. In each group, methods are categorized based on used modalities (RGB, depth, skeleton, or combination of multiple modalities). 

\subsection{Motion-based Feature Approaches}
The first category is based on pre-computed motion features like 2D dense optical flow maps as input to the neural networks. These networks generally use multiple streams to encode both appearance and motion of human actions using different modalities. Finally, different types of information (learned from the input) are fused to get the final result. There are different fusion approaches in the literature. In \cite{RN289}, fusion methods are categorized into early, late, and intermediate. Early fusion involves the integration of multiple raw or preprocessed data modalities into a vector ahead of feature extraction. In intermediate fusion, the features, respective to each stream, are concatenated before classification. Late fusion refers to collecting decisions from multiple classifiers and applying maximum or average scores to get the final decision. Similar taxonomies also exist in the literature; for example, in \cite{RN290} fusion methods are grouped into feature-level, score-level, and decision-level.
\subsubsection{RGB}
In \cite{RN73}, flow coding images computed from consecutive video frames are fed to a deep CNN network to extract deep temporal features from flow coding images. Then, the output features of several frames are concatenated together to learn the temporal convolution. Finally, a fully connected feedforward neural network is used for classification. In \cite{RN19}, two separate streams (a spatial and a temporal convolution network) are simultaneously applied to learn both the appearance and motion of actions. In \cite{RN74}, a trajectory-pooled deep-convolutional descriptor is proposed to learn discriminative convolutional feature maps, and conduct trajectory-constrained pooling to aggregate these convolutional features into effective descriptors. In \cite{RN75}, temporal linear encoding (TLE), embedded inside of ConvNet architectures is presented to aggregate information from an entire video. The pooling layer called ActionVLAD \cite{RN76} is also used to combine appearance and motion streams. It aggregates convolutional feature descriptors in different image portions and temporal spans. This layer is used to combine appearance and motion streams. As mentioned before, the problem with two-stream networks is the lack of transferring knowledge between the two streams \cite{RN19,RN74}. Some approaches address this problem \cite{RN77} to communicate between the two streams. However, this interaction between different streams is known difficult \cite{RN8}.
\subsubsection{Depth}
\begin{figure}
\centering
\includegraphics[width=0.98\columnwidth]{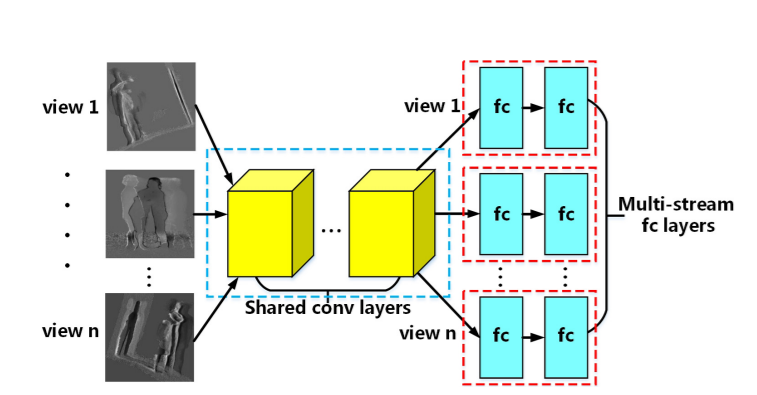}
\caption{Multi-view dynamic image adaptive CNN learning model \cite{RN78}.}
\label{fig:image2}
\end{figure}

In \cite{RN78}, a CNN-based framework is proposed using dynamic images. Dynamic images (DIs) summarize the motion and temporal information of video sequences in a single image. Multi-view dynamic images created from multi-view depth video are employed to better model 3D specifications.  Different views share the same convolutional layers but there are distinct fully connected layers (Figure \ref{fig:image2}). The main goal is to reduce the gradient vanishing problem, particularly on the shallow convolutional layers. Further, spatial-temporal action proposal is used to decrease the sensitivity of CNNs to scene variations.

In \cite{RN79}, three different depth representations are proposed: dynamic depth image (DDI), dynamic depth normal image (DDNI), and dynamic depth motion normal image (DDMNI) for segmented and continuous action recognition. Dynamic images are constructed using hierarchical bidirectional rank pooling to extract spatial-temporal information. DDIs extract dynamics of postures, while DDNIs and DDMNIs exploit 3D structural information of depth maps. These three representations of depth are fed to a pre-trained CNN model for fine-tuning without any need for training the network from scratch. In \cite{RN80}, weighted hierarchical depth motion maps (WHDMM) and three-channel deep convolutional neural networks (ConvNets) are suggested using depth maps of small human action recognition datasets. Depth maps from different viewpoints are used to make WHDMMs extract spatio-temporal features of actions into 2-D spatial structures. Then, these structures are converted to pseudo-color images. Finally, color-coded WHDMMs are trained via distinct pre-trained ConvNets. In \cite{RN81}, a method for human action recognition from depth sequences is presented. Firstly, to form the depth motion maps (DMMs), the raw frames are projected onto three orthogonal Cartesian planes and the results are stacked into three still images (corresponding to the front, side, and top views). Then, the local ternary pattern (LTP) is introduced as an image filter for DMMs to improve the distinguishability of similar actions. Finally, corresponding LTP-encoded images are classified using CNN.

\subsubsection{Skeleton}
In \cite{RN82} and \cite{RN83} spatio-temporal information carried in 3D skeleton sequences is represented by three 2D images, referred to as joint trajectory maps (JTM), through encoding the joint trajectories and their dynamics into color distribution in the images. Then ConvNets are adopted to learn the discriminative features for human action recognition. Such an image-based representation enables to use of existing ConvNets models for the classification of skeleton sequences without training the networks afresh. In \cite{RN84} and \cite{RN85}, a skeleton image representation named SkeleMotion is introduced to be used as input of CNNs. In this method, the temporal dynamics are encoded by explicitly computing the magnitude and orientation values of the skeleton joints. Different temporal scales are employed to compute motion values to aggregate temporal dynamics to the representation. In \cite{RN86}, the spatio-temporal information of a skeleton sequence is encoded into color texture images, called skeleton optical spectra. The encoding consists of four steps; mapping of joint distribution, spectrum coding of joint trajectories, spectrum coding of body parts, and joint velocity weighted saturation and brightness. Again, convolutional neural networks are used to learn the discriminative features for action recognition. In \cite{RN87}, color texture images referred to as joint distance maps (JDMs) along with ConvNets are employed to exploit the discriminative features from the JDMs for human action and interaction recognition. The pair-wise distances between joints over a sequence of single or multiple-person skeletons are encoded into color variations to capture temporal information. In \cite{RN88}, the 3D coordinates of the human body joints carried in skeleton sequences are transformed into image-based representations and stored as RGB images. Then a deep architecture based on ResNets is proposed to learn features from obtained color-based representations and classify them into action classes.

\subsubsection{RGB \& Depth}
In \cite{RN89}, Asadi-Aghbolaghi et al. investigated the combination of hand-crafted features and deep techniques in human action recognition via RGB-D videos. Multimodal dense trajectories (MMDT) are created from RGB, depth, scene flow, and optical flow modalities which are the inputs to 2DCNNs. Dynamic images such as depth motion maps and motion history images (MHI) are the other pre-computed motion features initially presented in \cite{RN90}. These features grounded on rank pooling summarize the motion and action information of a video in a single image to represent the whole sequence. A two-stream CNN network pre-trained on VGG16 \cite{RN91} or ResNet-101\cite{RN92} is suggested in \cite{RN93}. Dynamic images created independently from RGB videos and depth sequences are fed to the network. Finally, extracted features are concatenated and fed through a fully connected layer for action class prediction. Some studies suggest the difference of successive DMM frames projected on XY, YZ, and XZ planes corresponding to front, side, and top. Singh et al. \cite{RN94} employed dynamic images created from RGB videos and three DMMs as inputs to the pre-trained VGG-F model \cite{RN91}. A weighted product model is used to categorize the activity. Pre-trained networks with four streams are proposed in \cite{RN95,RN96} that accept an MHI created from RGB and DMMs in three distinct views (top, front, and side). The score of each stream is late fused at the end of the network to categorize the activity.  For surgical recognition tasks, Twin et al. \cite{RN97} suggest a four-stream CNN network pre-trained on AlexNet \cite{RN98} using RGB, depth, and their motions as DNN entries. Then, features are concatenated and the classifier predicts the class label.  In \cite{RN99}, four-channel data is used via combining RGB and depth, where extracting the scene flow from RGB-D videos is considered for action recognition to summarize the RGB-D videos. In this work, RGB and depth are considered as a single unit for extracting the features. 
A branch of studies tries to extract common-specific features of different modalities to increase the accuracy of action prediction. Combining the common and specific components in input features may be quite complicated and highly nonlinear. Shahroudy et al. \cite{RN100} proposed a deep shared-specific network using nonlinear autoencoder-based component factorization layers. Also, while RGB and depth images are inherently distinct in appearance, there is considerable consistency between them at a high-level \cite{RN101}, which will affect the classification accuracy. Qin et al. \cite{RN101} developed a unique two-stream framework to extract common-specific features through the constraint of similarity at a high level. In \cite{RN102}, a multi-stream deep neural network is suggested for egocentric action recognition. The work \cite{RN102} uses the complementary features of RGB and depth by learning the nonlinear structure of heterogeneous information. It strives to keep the unique features for each modality and concurrently explore their sharable information in a unified framework. In addition, it uses a Cauchy estimator to maximize the correlations of the sharable components and enforce the orthogonality constraints on the individual components to ensure their high independencies. The cross-modality complementary features are learned from RGB and depth modalities via a cross-modality compensation block (CMCB) \cite{RN103}. The CMCB initially extracts features from the two separate information flows, then sends and intensifies them to the RGB-D paths using the convolution layers. To increase action recognition performance, CMCB includes two general DNN architectures: ResNet and VGG. 
Wang et al. \cite{RN104} employ two distinct cooperative convolutional networks (c-ConvNet) to extract information from dynamic images comprised of both visual RGB (VDIs) and depth (DDIs).  The c-ConvNet comprises one feature extraction network and two branches, one for ranking loss and another for softmax loss. By utilizing bidirectional rank pooling, two dynamic images represent VDIs and DDIs: forward (f) and backward (b), VDIf \&VDIb and DDIf \& DDIb, respectively. In \cite{RN105}, segmented bidirectional rank pooling is used to gather spatio-temporal information. Moreover, the multimodality hierarchical fusion method gets the complementary information of multimodal data for categorization. The multimodality hierarchical system contains visual RGB and depth dynamic images, i.e., VDIs-f, VDIs-b, DDIs-f, and DDIs-b (f for forward and b for backward) and optical flow fields (X-stream and Y-stream) formed by ConvNets. Dynamic images are created from the RGB-D series as ConvNets entries to extract spatio-temporal information \cite{RN106}. Then a segmented cooperative ConvNet is applied to learn the complementary information of RGB-D modalities. In \cite{RN107}, RGB and depth frames are used as training inputs, but only RGB is employed during test time. A hallucination network is utilized to simulate the depth stream for test time. A strategy based on inter-stream connection is used to improve the hallucination network's learning process. A loss function that combines distillation and privileged information is also developed.

\subsubsection{RGB \& Skeleton}
Verma et al. \cite{RN108} proposed a two-stream framework to exploit spatio-temporal features using both CNNs and RNNs. Motion history image and motion energy image (MEI) are the RGB descriptors. Further, the skeleton modality is used after developing intensity images in three views: top, side, and front. Features of each stream are fused and the final prediction is performed based on scores of each stream using the weighted product rule.
Tomas et al. \cite{RN109} utilized appearance and motion information from RGB and skeleton joints to detect fine-grained motions. Motion representations are learned by CNN and motion history images that are generated from RGB images. In addition, stacked auto-encoders measure the distances of the joints from the mean joint in each frame to consider discriminative movements of human skeletal joints. 

\subsubsection{Depth \& Skeleton}
Kamel et al. \cite{RN110} employed depth motion image (DMI), moving joint descriptor (MJD), and fusion of DMI with MJD as inputs of the suggested CNN framework. DMI represents the body changes of depth maps in an image, while MJD indicates body joint position and orientation changes around a fixed point. Wang et al. \cite{RN111} utilized the bidirectional rank pooling approach to three hierarchical spatial levels of depth maps driven by skeletons; body, part, and joint. Each level featured various components, which possessed joint positions. Spatio-temporal and structural information at all levels is learned via a spatially structured dynamic depth image (S2DDI) conserving the coordination and synchronization of body parts throughout the action. Besides, this framework contains three weights-shared ConvNets and scored fusion for classification. In \cite{RN112},  a CNN-based human action recognition framework is proposed by fusing depth and skeleton modalities. The proposed adaptive multiscale depth motion maps (AM-DMMs) computed from depth maps capture shape and motion cues. Moreover, adaptive temporal windows help the robustness of AM-DMMs in front of motion speed variations. In addition, a method is also proposed for encoding the spatio-temporal information of each skeleton sequence into three maps, called stable joint distance maps (SJDMs) which describe spatial relationships between the joints. A multi-channel CNN is adopted to exploit the discriminative features from texture color images encoded from AM-DMMs and SJDMs for recognition. 

\subsubsection{RGB \& Depth \& Skeleton}
Singh et al. \cite{RN113} introduced a modality fusion technique called deep bottleneck multimodal feature fusion (D-BMFF) framework for three modalities of RGB, depth, and skeleton. 3D joints are transformed into a single RGB skeleton motion history image (RGB-SklMHI). Every ten RGB and depth frames with a single Skel-MHI image are fed to the framework to extract spatial and temporal features respectively.   Extracted features of three-modality streams are combined by multiset discriminant correlation analysis. Then action classification is performed using a linear multiclass SVM. Khaire et al. \cite{RN114} aim to enhance activity recognition by using skeleton images, a motion history image, and three depth motion maps from the side, top, and front as inputs of a five-stream CNN network. Elmadany et al. \cite{RN115} introduced two fusion approaches to exploit common subspace from two sets and more than two sets, i.e., biset globality locality preserving canonical correlation analysis (BGLPCCA) and multiset globality locality preserving canonical correlation analysis (MGLPCCA), respectively. These strategies represent global and local data features using low-dimensional shared subspace. Besides, two descriptors are suggested for skeleton data and depth. Finally, a framework composed of proposed fusion methods and descriptors is used for action recognition.  In \cite{RN116}, various rank pooling and skeleton optical spectra approaches are examined to create dynamic images from RGB-D and skeleton. Dynamic images are divided into five categories: a dynamic color group (DC), a dynamic depth group (DD), and three dynamic skeleton groups (DXY, DYZ, DXZ). Several dynamic images featuring the major postures for each group are developed to represent different action postures. Then, a pre-trained flow-CNN extracting spatio-temporal features are used with a max-mean aggregation. 
Wu et al. \cite{RN117} described a deep hierarchical dynamic neural network for gesture recognition. The suggested framework is composed of a Gaussian-Bernouilli deep belief network (DBN) to extract dynamic skeletal features and a 3DCNN to represent features from RGB and depth images. Furthermore, intermediate and late fusion techniques are used to fuse RGB and depth with the skeleton. Finally, HMM predicts the gesture class label by learning emission probabilities. Romaissa et al. \cite{RN118} proposed a four-step framework for action recognition. First dynamic images are created from RGB-D videos, and features of dynamic images are extracted via a pre-trained model utilizing the transfer learning approach. Then, the Canonical correlation analysis method fuses extracted features. Finally, a bidirectional LSTM is trained to recognize action labels.

\subsubsection{Discussions }
In brief, utilizing motion-based features is a common approach for modeling temporal variations using distinct or multiple data modalities that allows using pre-trained 2D ConvNets for modeling motion information. These networks generally exploit multiple streams of convolutional networks to encode both appearance and motion of human actions using different modalities. Descriptors such as motion history/energy images for RGB, dynamic images for depth, and joint trajectory maps for skeleton are popular pre-computed motion features used for human action recognition. The most important problem in multi-stream networks is the necessity of communications between different streams to transfer information in learning multimodal spatiotemporal features. The lack of effective interactions between the streams is one of the major problems in multi-stream networks. Such interactions are important for learning spatiotemporal features. Finally, the multi-stream CNN architectures learn different types of information from the input (through separate networks) and then perform fusion to get the result. This enables the traditional 2D CNNs to effectively handle the video data and achieve high accuracy. However, this type of architecture is not powerful enough for modeling long-term dependencies, i.e., it has limitations in effectively modeling the video-level temporal information.

\subsection{Three-Dimensional Filters}
In the second category, spatiotemporal filters (for example 3D convolution and 3D pooling) are used in the convolutional layer. Convolution as the essential operation in CNNs calculates pixel values according to a small neighborhood using a kernel (filter). Spatio-temporal filters extend 2D convolution networks by using 3D convolution. 3D convolution captures the temporal dynamics over some successive frames. However, some approaches convert the entire sequence into a 2D image and then use conventional 2D filters.

\subsubsection{RGB}
In \cite{RN119}, the effects of different spatiotemporal convolutions are studied for action recognition. In this work, the improvement of the accuracy of 3D CNNs over 2D CNNs is empirically demonstrated within the framework of residual learning. In \cite{RN120}, FAST 3D convolutions are introduced, a convolution block that combines a 2D spatial convolution with two orthogonal spatio-temporal convolutions. The block is motivated by the often characteristic horizontal and vertical motion of human actions. In \cite{RN17,RN18,RN121,RN122,RN123}, the 3D convolution over consecutive frames is applied for action recognition. In \cite{RN18}, the developed deep architecture model produces multiple channels of information from adjacent input frames and performs convolution and subsampling separately in each channel. Finally, feature representation is obtained by aggregating information from all channels. In \cite{RN122}, pseudo-3D residual net architecture is proposed which aims to learn spatio-temporal video representation in deep networks by simplifying 3D convolutions with 2D ﬁlters on spatial dimension plus 1D temporal connections. In \cite{RN123}, the learning of long-term video representations is considered by studying architectures with long-term temporal convolutions (LTC). To keep the complexity of networks tractable, the temporal extent of representations is increased at the cost of decreased spatial resolution.

\subsubsection{Depth}
In \cite{RN124}, a 3D full CNN-based framework, called 3DFCNN, is developed for real-time human action recognition from depth videos captured from an RGB-D camera. The network exploits spatio-temporal information of depth sequences to use in the categorization of actions. The aim is the use of depth in privacy-aware systems because people’s identities are not recognized from depth images.

\subsubsection{Skeleton}
In \cite{RN125}, a multi-scale temporal modeling module is designed following \cite{RN126}. This module contains four branches, each containing a 1 × 1 convolution to reduce channel dimension. The first three branches contain two temporal convolutions with different dilations and one Max-Pool respectively following 1 × 1 convolution. The results of the four branches are concatenated to obtain the output. In \cite{RN127}, a pre-trained 2D convolutional neural network is used as a pose module. A pre-trained 3DCNN is also used as an infrared module to respectively extract features from skeleton data and visual features from videos. Both feature vectors are then fused and jointly classified using a multilayer perceptron (MLP).

\subsubsection{RGB \& Depth}
Li et al. \cite{RN128} employed RGB and depth as inputs to a pre-trained C3D network for gesture recognition. Extracted features are concatenated or averaged. Finally, the framework uses a linear SVM as a classifier.  Zhu et al. \cite{RN129} employed pyramid input and fusion with multiscale contextual information via 3D CNNs to learn gestures from the whole video. Zhang et al. \cite{RN130} presented 3D lightweight structures for action recognition based on RGB-D data. The suggested lightweight 3D CNNs have considerably fewer parameters with reduced computing costs, and it results in desired recognition performance compared to common 3D CNNs. 

Qin et al. \cite{RN131} employed 3D CNNs to extract common-specific features from RGB-D data. A novel end-to-end trainable framework called TSN-3DCSF is proposed for this purpose. In \cite{RN132}, a fusion approach is proposed using the adaptive cross-modal weighting (ACmW) approach to extract complementarity features from RGB-D data. ACmW block explores the relationship between the complementary information from multiple streams and fuses them in the spatial and temporal dimensions. In \cite{RN133}, a regional attention with architecture-rebuilt 3D network (RAAR3DNet) is suggested for gesture recognition. Fixed Inception modules are replaced with the automatically rebuilt structure through neural architecture search (NAS) to acquire the varied representations of features in the early, middle, and late levels of the network. In addition, a stacking regional attention module called dynamic-static attention (DSA) is used to highlight the hand/arm regions and the motion information. 

\subsubsection{Depth \& Skeleton}
Liu et al. \cite{RN134} suggest a 3D-based deep convolutional neural network (3D2CNN) to learn depth features along with the joint vector containing skeletal features. Finally, the decision fusion of SVM classifiers demonstrates the action class. 

\subsubsection{Discussions }
Shortly, 3D filters as the extension of 2D filters capture the temporal dynamics at the cost of requiring more parameters than 2D convolution networks. The advantage is capturing discriminative features along both spatial and temporal dimensions while the disadvantage is the limitation to a certain temporal structure (by considering very short temporal intervals) and complicated encoding of long-term temporal data. The 3D CNN-based methods generally perform spatio-temporal processing over limited intervals (using the window-based 3D convolutional operations), where each convolutional operation is only applied to a relatively short-term context in videos. For multi-modal approaches, the 3D filters can be applied to distinct modalities to simultaneously capture spatial and temporal intra-modal features. For this purpose, multi-stream networks and different strategies for fusion may be applied. However, the fusion of features is again a concern. Score fusion and feature fusion are two widely used multi-modality fusion schemes in human action recognition \cite{RN45}. The score fusion integrates the separately made decisions based on different modalities to produce the final results. Meanwhile, feature fusion generally combines the features from different modalities to yield aggregated and powerful features for recognizing different actions. However, existing multi-modality methods are not as effective as expected owing to a series of challenges, such as over-fitting \cite{RN45,RN135}.

\subsection{Temporal Sequence Models}
The third group usually aggregates CNN features applied at individual frames with temporal sequence models such as recurrent neural networks \cite{RN136}. RNNs that are designed to work with sequential data, use the previous information in the sequence to produce the current output. The main problem with RNNs is the short-term memory problem, caused by the vanishing gradient problem. As RNN processes more steps, it suffers from vanishing gradient more than other neural network architectures. To overcome this problem, two specialized versions of RNNs are created; GRU (gated recurrent unit)\cite{RN137} and LSTM (long short-term memory) \cite{RN138}. LSTM and GRU use memory cells to store the information of previous data in long sequences using gates. Gates that control the flow of information in the network, are capable of learning the importance of inputs in the sequence and storing or passing their information in long sequences. GRU structure is less complex compared with LSTM because it has less number of gates (two gates of reset and update for GRU compared with three gates of input, output, and forget for LSTM). However, other temporal sequence models like the hidden Markov model (HMM) are also applied \cite{RN117} in the literature together with CNN features.

\subsubsection{RGB}
In \cite{RN120,RN139}, both CNNs and LSTMs are utilized for capturing spatial motion patterns along with temporal dependencies. In \cite{RN140}, a conflux long short-term memory network is proposed to recognize actions from multi-view cameras. The proposed framework first extracts deep features from a sequence of frames using a pre-trained VGG19 CNN model for each view. Second, the extracted features are forwarded to the conflux LSTM network to learn the view of self-reliant patterns. In the next step, the inter-view correlations using the pairwise dot product are computed from the output of the LSTM network corresponding to different views to learn the view inter-reliant patterns. Finally, flattened layers followed by a softmax classifier are used for action recognition.

\subsubsection{Depth}
In \cite{RN141}, two networks based on ConvLSTM are suggested with different learning strategies and architectures.  One network uses a video-length adaptive input data generator (stateless) while the latter discovers the stateful capability of general recurrent neural networks, but is applied in the specific case of human action recognition. This property allows the model to gather discriminative patterns from previous frames without compromising computer memory. In \cite{RN142}, the ConvLSTM network is used with depth videos for home caring of elderly adults. In \cite{RN143}, a bidirectional recurrent neural network (BRNN) is developed for depth-based human action recognition. First, the 3D depth image is projected on three 2D planes and is fed to three distinct BRNNs. In the following layers, extracted features of each BRNN are fused and fed to the next BRNNs. The network follows with fully connected and softmax layers.

\subsubsection{Skeleton}
In \cite{RN144}, an end-to-end trainable hierarchical RNN model is developed using skeleton data for recognizing activities. The human skeleton is divided into five body parts instead of the whole skeleton data in the training phase. Then, each part is separately fed to a subnet. Next, extracted features by each subnet are hierarchically fused and fed to the higher layer. In the end, a high-level representation of the skeleton is used for the final classification. In \cite{RN145,RN146}, a universal spatial RNN-based model uses geometric features. The multi-stream LSTM network is trained with different geometric features and a new smoothed score fusion method is used. The potential of learning complex time-series representations via high-order derivatives of states is investigated in \cite{RN147}. In this work, a differential gating scheme is proposed for the LSTM neural network to highlight the change in information gain due to salient motions between consecutive frames. The proposed differential recurrent neural network (dRNN) quantifies the change in information gained by the derivative of states. In \cite{RN148}, an end-to-end fully connected deep LSTM framework is proposed for action recognition. The co-occurrences of skeleton joints are learned via a regularization mechanism. Further, a dropout algorithm is suggested for gates, cells, and output responses of neurons. In \cite{RN149}, RNNs are also used to model the temporal dependencies of the features of body parts in actions. In \cite{RN150}, a three-structure-based traversal method is proposed. Besides, a new gating scheme in LSTM is proposed to handle noise and occlusion of skeleton data that learns the reliability of the sequential input data and adjusts its effect on updating the long-term context information stored in the memory cell. A two-stream RNN architecture is suggested to model spatial and temporal features of actions with skeleton data as input \cite{RN151}. Two different structures are designed for the temporal stream, including stacked RNN and hierarchical RNN. Further, spatial structure is modeled by two methods. In addition, 3D-based data augmentation techniques such as rotation and scaling transformation are suggested. In \cite{RN152}, an ensemble temporal sliding LSTM (TS-LSTM) network is introduced for skeleton-based action recognition, which consists of several parts including short-term, medium-term, and long-term TS-LSTM networks. Then, with an average ensemble among different parts various temporal dependencies are captured. In addition, features of multiple parts are visualized to demonstrate the relation between recognized action and its correspondent multi-term TS-LSTM features. In \cite{RN153}, it is suggested to use an independently recurrent neural network (IndRNN). Besides, network weights are regularized to resolve the gradient vanishing problem. Moreover, IndRNN is over ten times faster than the commonly used LSTM. In \cite{RN154}, an attentional recurrent relational network-LSTM (ARRN-LSTM) is proposed that models spatial and temporal dynamics in skeletons for action recognition. The recurrent relational part of the network learns the spatial features of a single skeleton, followed by a multi-layer LSTM that learns the temporal features in the skeleton sequences. An adaptive attentional module is used between the two modules to focus on the most discriminative parts in the single skeleton. In addition, a two-stream architecture is used to learn the structural features among joints and lines to use the complementarity from different geometries in the skeleton.

\subsubsection{RGB \& Depth}
Pigou et al. \cite{RN155} developed an end-to-end trainable network employing temporal convolutions and bidirectional recurrence. RGB and depth are considered as four-channel data or a 4D entity.  In this method, RNNs represent high-level spatial information. In addition, RNNs predict the beginning and ending frames of gestures. In \cite{RN156}, two-stream RNNs are used for gesture recognition that utilizes RGB-D data to represent the contextual information of temporal sequences. 

\subsubsection{Depth \& Skeleton}
Mahmud et al. \cite{RN157} employed depth quantized images and skeleton joints for dynamic hand gesture recognition. Both CNN and LSTM structures are used in the network to extract depth features, while skeleton features are extracted via LSTM following distinct MLPs. Fused scores are used in prediction with MLP scores of fused extracted features from quantized images and skeleton joints in the previous process. Lai et al. \cite{RN158} proposed a framework composed of CNNs and RNNs using depth and skeleton to hand gesture recognition. Further, several fusion strategies were investigated for enhancing performance, including feature-level fusion and score-level fusion.  Shi et al. \cite{RN159} proposed a privileged information-based recurrent neural network (PRNN). The privileged information (PI) is only provided during training but not through the testing procedure. This model considered skeletal joints as a PI in three-phase training processes, including; pre-training, learning, and refining. The recommended network was end-to-end trainable and the CNN and RNN parameters were jointly acquired. The final network enhances latent PI iteratively in an EM procedure.

\subsubsection{RGB \& Depth \& Skeleton}
Hu et al. \cite{RN160} proposed a framework to learn modality-temporal mutual information from tensors called the deep bilinear framework. The bilinear block learns the time-varying dynamics and multimodal information consists of modality pooling and temporal layers. The deep bilinear model is created through accumulating bilinear blocks and other layers to extract video modality-temporal information. Further modality-temporal cube descriptor is presented as deep bilinear learning input. 

\subsubsection{Discussions }
Finally, another common approach in modeling temporal variations is using the features applied at individual frames with temporal sequence models such as RNNs. Especially for RGB and depth, it is straightforward to use CNNs to extract spatial information and then use RNNs to extract temporal information of a sequence. Although the third group can deal with longer-range temporal relations, temporal sequence models such as RNN or LSTM can only exploit partial temporal information because regular RNNs cannot access all input elements at each given time step. Having access to all elements of a sequence at each time step can be overwhelming. To help the RNNs focus on the most relevant elements, the attention mechanism can be used that assigns different attention weights to each input element. Since the skeleton encodes high-level information about important details of a scene, the skeleton may be used to guide RGB/depth features. So, the important information strongly related to the action is enhanced.

\subsection{Transformers and Attention}
Before the arrival of transformers, most state-of-the-art methods were based on gated RNNs (such as LSTMs and GRUs) with considering attention mechanisms. Recently, transformer models such as BERT (bidirectional encoder representations from transformers) \cite{RN161}, GPT (generative pre-trained transformer) \cite{RN162}, RoBERTa (robustly optimized BERT pre-training) \cite{RN163}, and T5 (text-to-text transfer transformer) \cite{RN164} has shown promising results in the field of NLP for tasks such as text classification and translation \cite{RN25,RN165}.  Following these results, transformers are starting to be used in the ﬁeld of computer vision (which was dependent on deep ConvNets and RNNs in the last decade) by introducing models such as ViT \cite{RN26} and DeiT \cite{RN166} for image classiﬁcation, DETR for object detection \cite{RN167}, and VisTR for video instance segmentation \cite{RN168}. 
For action recognition, considering the sequential nature of video makes it a perfect match for transformers to be used for modeling temporal variations. Although the application of transformers to action recognition is relatively new, the amount of research that has been proposed on this topic within the last few years is surprising. Now, transformers are built on attention technologies without using a recurrent neural network backbone, demonstrating the ability of the attention mechanisms alone compared with RNNs along with attention. Some approaches are strictly dependent on the transformer and self-attention mechanisms \cite{RN169} to extract spatio-temporal features. Some others use CNN features besides transformers to make benefit from both architectures \cite{RN170,RN171,RN172}.

\subsubsection{RGB}
\begin{figure}
\centering
\includegraphics[width=0.98\columnwidth]{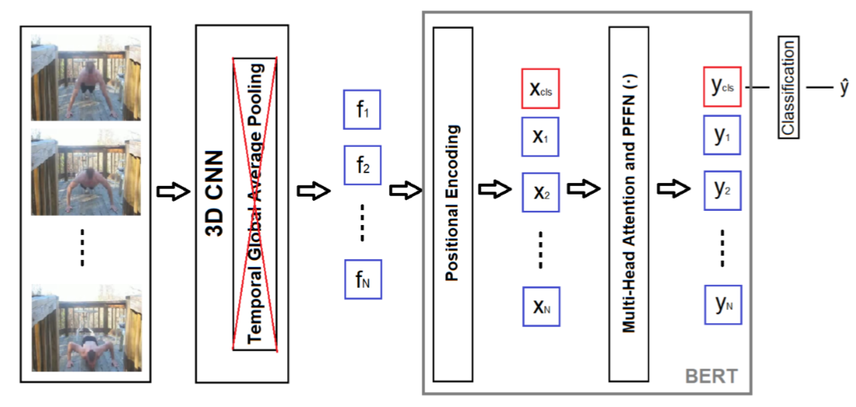}
\caption{BERT-based Temporal Modeling with 3D CNNs for Action Recognition \cite{RN184}.}
\label{fig:image3}
\end{figure}

In \cite{RN173}, an attention-based model for action recognition is proposed. The model can selectively concentrate on important elements in video frames and dynamically pool convolutional feature maps to produce discriminative features using long short-term memory units. In \cite{RN174}, hierarchical RNN and attention mechanisms are applied to capture both short-term and long-term motion information. In \cite{RN175}, Girdhar and Ramanan proposed a new method for approximating bilinear pooling with low-rank decomposition. This yields an attentional pooling that substitutes calculating the second-order features with the product of two attention maps of top-down and bottom-up attention maps. Li et al. \cite{RN176} introduced motion-based attention along with LSTM for end-to-end sequence learning of actions in the video.  In \cite{RN177}, Du et al. proposed a spatio-temporal attention mechanism to selectively focus on spatial visual elements as well as keyframes. In \cite{RN178}, the spatial transformer network \cite{RN179} is introduced with an attention mechanism to explicitly model the spatial structures of human poses. In \cite{RN180}, a convolutional LSTM algorithm based on the attention mechanism is proposed to improve the accuracy of action recognition by mining the salient regions of actions in videos. First, GoogleNet \cite{RN181} is used to extract the features of video frames. Then, those feature maps are processed by the spatial transformer network for attention. Finally, to classify the action, the sequential information of the features is handled by the convolutional LSTM network. In \cite{RN182}, a video action recognition network called action transformer is proposed that uses a modiﬁed transformer architecture as a 'head' to classify the action of a person of interest. It combines two other ideas of using a spatiotemporal I3D model \cite{RN121} as the base backbone to extract features and a region proposal network (RPN) \cite{RN183} to localize people performing actions. The I3D features and RPN produce the query that is the input for the transformer head and combines contextual information from other people and objects in the surrounding video. In this way, the network can implicitly learn both to track distinct persons and to consider the actions of other people in the video. In addition, the transformer attends to the hand and face as the most reassuring parts when discriminating an action. In \cite{RN184}, 3D convolution is combined with late temporal modeling for action recognition. For this purpose, the temporal global average pooling (TGAP) layer at the end of 3D convolutional architecture is replaced with the BERT layer to model the temporal information with BERT’s attention mechanism (see Figure \ref{fig:image3}). It was shown that this replacement improves the performances of popular 3D convolution architectures such as ResNet, I3D, SlowFast, and R(2+1)D for action recognition.

In \cite{RN171}, a sparse transformer-based Siamese network (called TBSN) is proposed for few-shot action recognition, which aims to recognize new categories with only a few labeled samples. TBSN applies the sparse transformer to learn the correlation and importance of video clips, and a new measurement to calculate the distance between samples. In this paper, an embedding module is designed based on sparse-transformer whose main ideas are attention mechanism and feedforward network. This method also substitutes the softmax function with the sparsemax function, where the sparsemax function can output zero probabilities. By introducing the sparsemax, zero attention values are assigned to clips containing noises. In \cite{RN185}, video transformer network (VTN) architecture is proposed for real-time action recognition. The VTN is made up of an encoder that processes each frame of input sequence independently with 2D CNN (ResNet-34 \cite{RN186}), and the decoder that integrates intra-frame temporal information in a fully-attentional feed-forward approach. In \cite{RN172}, two spatio-temporal feature extraction (GSF) and aggregation (XViT) modules are developed for action recognition: GSF is a spatio-temporal feature extracting module that can be plugged into 2D CNNs. XViT is a video feature extractor based on the transformer. The proposed method uses an ensemble of GSF and the XViT models to generate the final scores. In \cite{RN187}, a simple fully self-attentional architecture called action transformer (AcT) is introduced that exploits 2D pose representations over small temporal windows. In \cite{RN188}, a pure-transformer architecture adapted from the Swin transformer for image recognition \cite{RN189} is proposed for video recognition that is based on spatiotemporal locality inductive bias. So this model is supposed to be able to leverage the power of the pre-trained image models. The Swin transformer \cite{RN28} introduced the inductive biases of locality, hierarchy, and translation invariance and can be served as a general-purpose backbone for various image recognition tasks. In \cite{RN190}, a two-pathway transformer network (TTN) is proposed that uses memory-based attention to explicitly model the relationship between appearance and motion. Speciﬁcally, each pathway is designed to produce spatial appearance information or temporal motion information. Then the generated features from two pathways are combined at the end of the framework. Here a transformer-based decoder is used to capture the underlying relationship between the appearance and motion information to improve action recognition. The decoder takes different features as its query, key, and value inputs so that the transformer heads can aggregate contextual information from one modality’s features in the value input to update the other modality’s features in the query input.

\begin{figure}
\centering
\includegraphics[width=0.98\columnwidth]{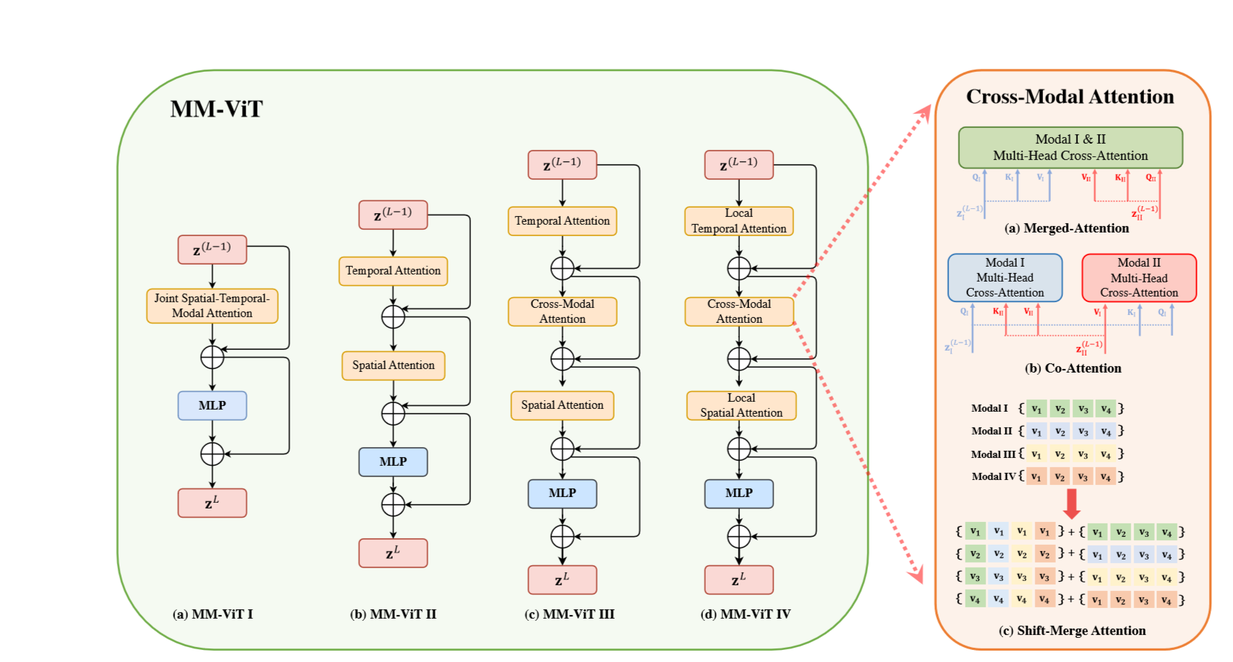}
\caption{Self-attention blocks in MM-ViT and cross-modal attention mechanisms \cite{RN169}.}
\label{fig:image4}
\end{figure}

In \cite{RN191}, TimeSFormer is proposed that extends ViTs to videos. In this method, the video is considered as a sequence of patches extracted from individual frames. In addition, to capture spatio-temporal relationships, divided attention is proposed to separately apply spatial and temporal attention within each block. In \cite{RN192}, multiscale vision transformers (MViT) are proposed for video and image recognition, to relate multiscale feature hierarchies with the transformer model. MViT hierarchically expands the feature complexity while reducing visual resolution. In \cite{RN193}, ViViT is proposed as a video vision transformer to extract spatio-temporal tokens from the input video, which are then encoded by a series of transformer layers. To handle the long sequences of tokens encountered in the video, several variants of the model which factorize the spatial and temporal dimensions of the input are proposed. In \cite{RN169}, a pure transformer-based approach called the multi-modal video transformer (MM-ViT) is proposed for video action recognition in the compressed video domain. MM-ViT exploits different modalities such as appearance (I-frames), motion (motion vectors and residuals), and audio waveform. To handle the large number of spatiotemporal tokens extracted from multiple modalities, four multi-modal video transformer architectures are introduced (see Figure \ref{fig:image4}). The simple architecture adopts the standard self-attention mechanism to measure all pairwise token relations. Three efﬁcient model variants are also presented with different approaches which factorize the self-attention calculation over the space, time, and modality dimensions. In addition, to explore the inter-modal interactions, three distinct cross-modal attention mechanisms are developed that can be integrated into the transformer architecture. Experimental experiments on public datasets demonstrate that MM-ViT performs better or equally well to the state-of-the-art CNN counterparts with much less computational cost (compared with the cost of optical flow).

\subsubsection{Skeleton}
\begin{figure}
\centering
\includegraphics[width=0.98\columnwidth]{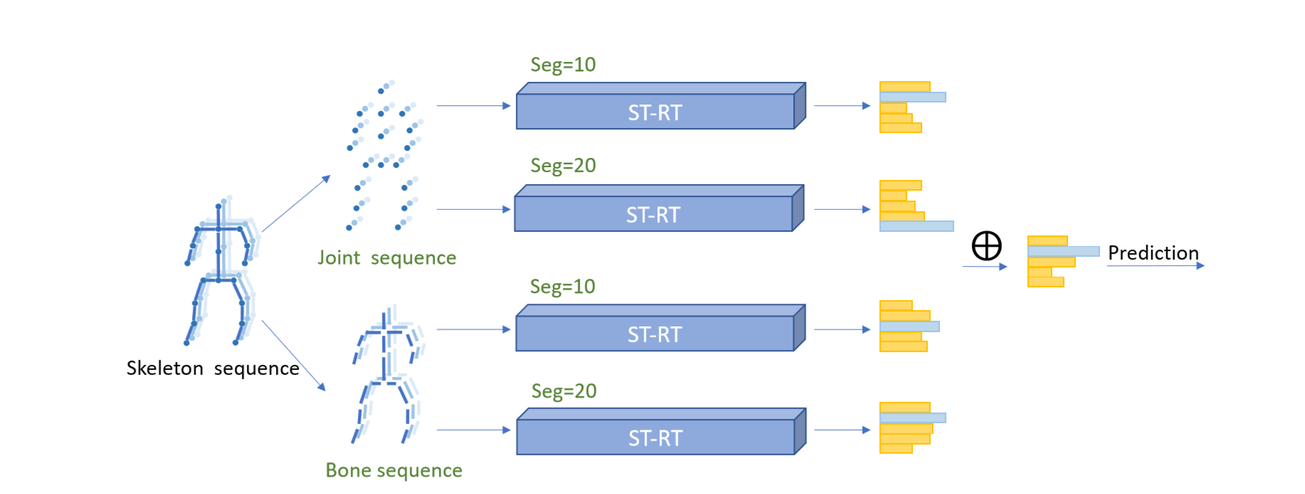}
\caption{The overall architecture of MSST-RT in \cite{RN200}.}
\label{fig:image5}
\end{figure}

For skeleton-based action recognition, some approaches use attention mechanisms for modeling dependencies. In \cite{RN194}, a global context-aware attention LSTM (GCA-LSTM) framework is suggested for action recognition to selectively focus on the informative joints of each frame. Further, a recurrent attention mechanism is proposed to enhance attention efficiency. Thereby, the proposed two-stream framework consists of coarse-grained and fine-grained attention. In \cite{RN195}, an LSTM-based approach called global context-aware attention LSTM (GCA-LSTM) is proposed to selectively focus on the informative joints in the action sequence using global contextual information. Besides, a recurrent attention mechanism for the GCA-LSTM network is introduced to achieve a reliable attention scheme for the action. An end-to-end spatial and temporal attention network is suggested for human action recognition from skeleton data in \cite{RN196}. The network is based on LSTM to learn selectively focus on discriminative joints in each frame, giving each frame a different degree of attention. In \cite{RN197}, an architecture named graph convolutional skeleton transformer (GCsT) is proposed to capture the long-term temporal context and enhance the flexibility of feature extraction for skeleton-based action recognition. The overall architecture is divided into three stages and each stage consists of two blocks. A spatial–temporal graph convolutional block (STGC) to extract the local-neighborhood relations and a spatial–temporal transformer block (STT) to capture global space–time dependencies. GCsT employs the benefits of both transformer and graph convolution network (GCN). In this way, hierarchy and local topology structure is conducted through GCN and the temporal attention and global context is provided with the transformer. In \cite{RN198}, a hierarchical transformer-based framework is proposed for modeling the spatio-temporal structure of a sequence of 3D human skeletons of human action. Speciﬁcally in this method, the 3D human skeletons are split into ﬁve human body parts, then they are fused hierarchically with self-attention layers based on the articulation of the human body parts. Besides, to predict the motion of the 3D skeleton it tries to model the body parts’ interactions and the motion directions. In \cite{RN199}, a transformer-based model called Motion-Transformer is proposed to capture the temporal dependencies via self-supervised pre-training on the sequence of human action. Besides, a flow prediction task is also introduced to pre-train the Motion-Transformer to capture the intrinsic temporal dependencies. The pre-trained model is then fine-tuned on the task of action recognition. In \cite{RN200}, a multi-stream spatial-temporal relative transformer architecture is also used instead of graph convolution or recurrence and LSTM, to capture long-range dependencies. The proposed architecture called relative transformer is based on standard transformer. The relative transformer module respectively evolves into a spatial relative transformer and temporal relative transformer to extract spatio-temporal features (ST-RT module). In addition, the dynamic representation module combines multi-scale motion information to handle actions with different durations. Lastly, four streams of ST-RTs modules with four dynamic data streams are combined to improve the performance (see Figure \ref{fig:image5}) where each stream extracts features from a corresponding skeleton sequence to complement each other.

In \cite{RN201,RN202}, a transformer self-attention approach is introduced in skeleton activity recognition as an alternative to graph convolution. This approach tries to model interactions between joints using a spatial self-attention module (SSA) to understand intra-frame interactions between diﬀerent body parts and a temporal self-attention module (TSA) to model inter-frame correlations. The two modules are combined in a two-stream network to produce the final score for action recognition. In \cite{RN203}, synchronous local \& nonlocal along with frequency attention (SLnL-rFA) model is proposed to extract synchronous detailed and semantic information from multi-domains. SLnL-rFA includes SLnL blocks for spatio-temporal learning and a residual rFA block for extracting frequency patterns. In \cite{RN204}, spatial transformer block and directional temporal transformer block are designed for modeling skeleton sequences in spatial and temporal dimensions respectively. To adapt to the imperfect information condition (due to occlusion, noise, etc.), a multi-task self-supervised learning method is also introduced by providing confusing samples in different situations to improve the robustness of the model. In \cite{RN205}, a transformer-based model is proposed with sparse attention and segmented linear attention mechanisms applied on spatial and temporal dimensions of action skeleton sequence to replace graph convolution operations with self-attention operations while requiring signiﬁcantly less computational and memory resources.

\subsubsection{RGB \& Skeleton}
For multimodal action recognition, the cross-modality features are also a concern besides spatio-temporal features. In \cite{RN206}, a spatio-temporal attention-based mechanism is proposed for human action recognition to automatically attend to the most important human hands and detect the most discriminative moments in an action. Attention is handled using a recurrent neural network and is fully differentiable. In contrast to standard soft-attention-based mechanisms, this approach does not use the hidden RNN state as input to the attention model. Instead, attention distributions are drawn using a human articulated pose as external information.  In \cite{RN207}, a closely related approach is proposed. However, the attention mechanism on the RGB space is conditioned on end-to-end learned deep features from the pose modality and not only hand-crafted pose features. In \cite{RN208}, an end-to-end network for human activity recognition is proposed leveraging spatial attention on human body parts. This paper proposes an RNN attention mechanism to obtain an attention vector for soft assigning different importance to human body parts using spatio-temporal evolution of the human skeleton joints. It also designs the joint training strategy to efﬁciently combine the spatial attention model with the spatio-temporal video representation by formulating a regularized cross-entropy loss to achieve fast convergence. 
In \cite{RN209}, an attention-based body pose encoding is proposed for human activity recognition. To achieve this encoding, the approach exploits a spatial stream to encode the spatial relationship between various body joints at each time point to learn the spatial structure of different body joints. In addition, it also uses a temporal stream to learn the temporal variation of individual body joints over the entire sequence. Later, these two pose streams are fused with a multi-head attention mechanism. It also captures the contextual information from the RGB video stream using an Inception-ResNet-V2 model combined with multi-head attention and a bidirectional long short-term memory network. Finally, the RGB video stream is combined with the fused body pose stream to give an end-to-end deep model for human activity recognition.

\subsubsection{RGB \& Depth}
\begin{figure}
\centering
\includegraphics[width=0.98\columnwidth]{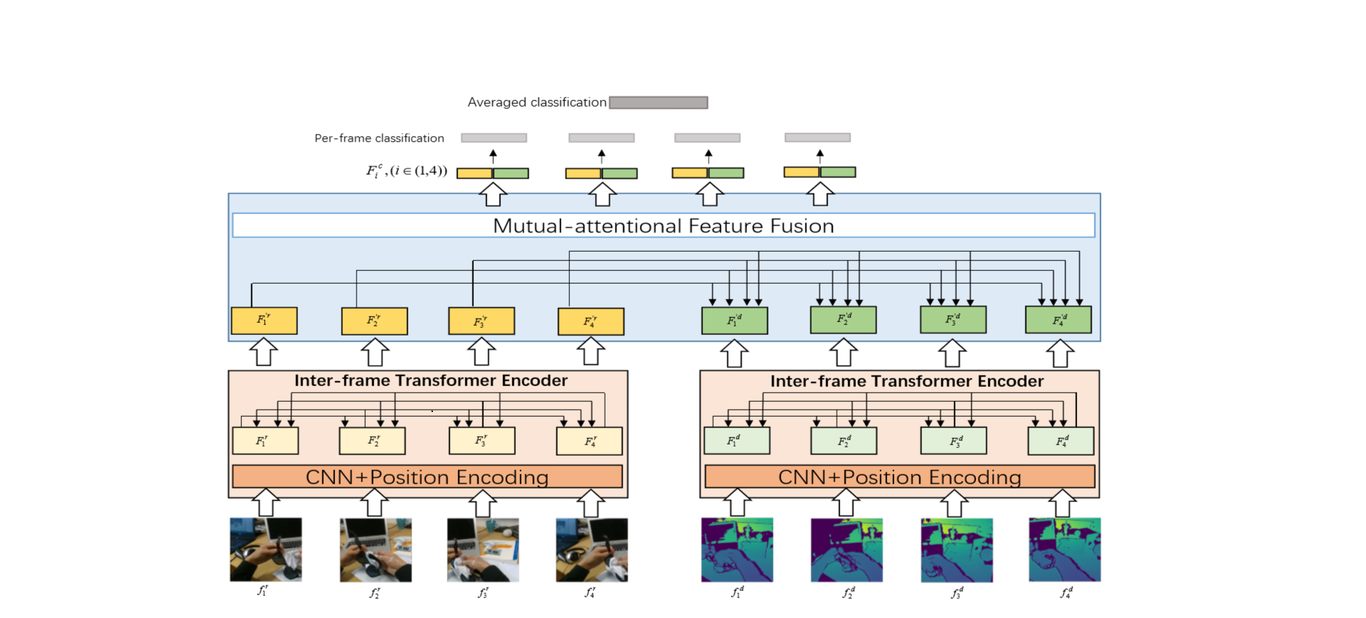}
\caption{proposed framework in \cite{RN210}. Features from each modality are interacted with and incorporated through the mutual-attentional block.}
\label{fig:image6}
\end{figure}

In \cite{RN210}, a transformer-based framework is proposed for egocentric RGB-D action recognition. It consists of two inter-frame transformer encoders and the mutual-attentional cross-modality modules (see Figure \ref{fig:image6}). The temporal information of distinct modalities is encoded through the self-attention mechanism. Then features from different modalities are fused via the mutual-attention layer. The inputs of this network are aligned RGB frames and depth maps (two streams). The frames are passed through a CNN and after average pooling are converted into two sequences of feature embeddings. Then both sequence features are fed to the transformer encoders to model the temporal structure respectively. Features obtained from the encoders interact through the cross-modality block and are fused to produce the cross-modality representation. The features are processed through the linear layer to get per-frame classiﬁcation. The final classification is performed by averaging the decisions over the frames of the video. 

\subsubsection{Discussions }
From studied papers, the interest in using attention-based and transformer networks for human action recognition is growing. These networks rely on self-attention mechanism to model dependencies across features over time, so the network can selectively extract the most relevant information and relationships. In addition, the great advantage of purely transformer-based networks is the fast-learning speed, and the lack of sequential operation, as with recurrent neural networks. Although video transformers have achieved promising results, they suffer from severe memory and computational overhead \cite{RN45}. In addition, to obtain the input to the transformer, a video is mapped to a sequence of tokens and then the positional embedding is added. A straightforward method of tokenizing the input video \cite{RN193} is to uniformly sample frames from the input video clip, embed each 2D frame independently using the same method as ViT \cite{RN26}, and concatenate all these tokens together. However, this method results in a large number of tokens which increases the computation. Attention can also be guided through different modalities. In the case of multimodality, transformers are used for intra-modality spatial and temporal modeling and cross-modality feature fusion. Handling a large number of spatiotemporal tokens extracted from multiple modalities is a concern.

\subsection{Hybrid Methods}
\begin{figure}
\centering
\includegraphics[width=0.98\columnwidth]{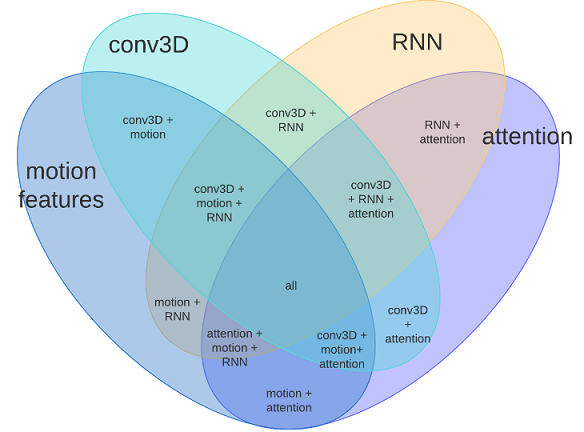}
\caption{Different Combinations of the aforementioned approaches.}
\label{fig:image7}
\end{figure}

In a group of studies, combinations of two or more techniques are used to exploit spatiotemporal dynamics. Considering the four aforementioned approaches, 11 different arrangements are probable for the hybrid techniques (see Figure \ref{fig:image7}).
\subsubsection{RGB}
\begin{figure}
\centering
\includegraphics[width=0.98\columnwidth]{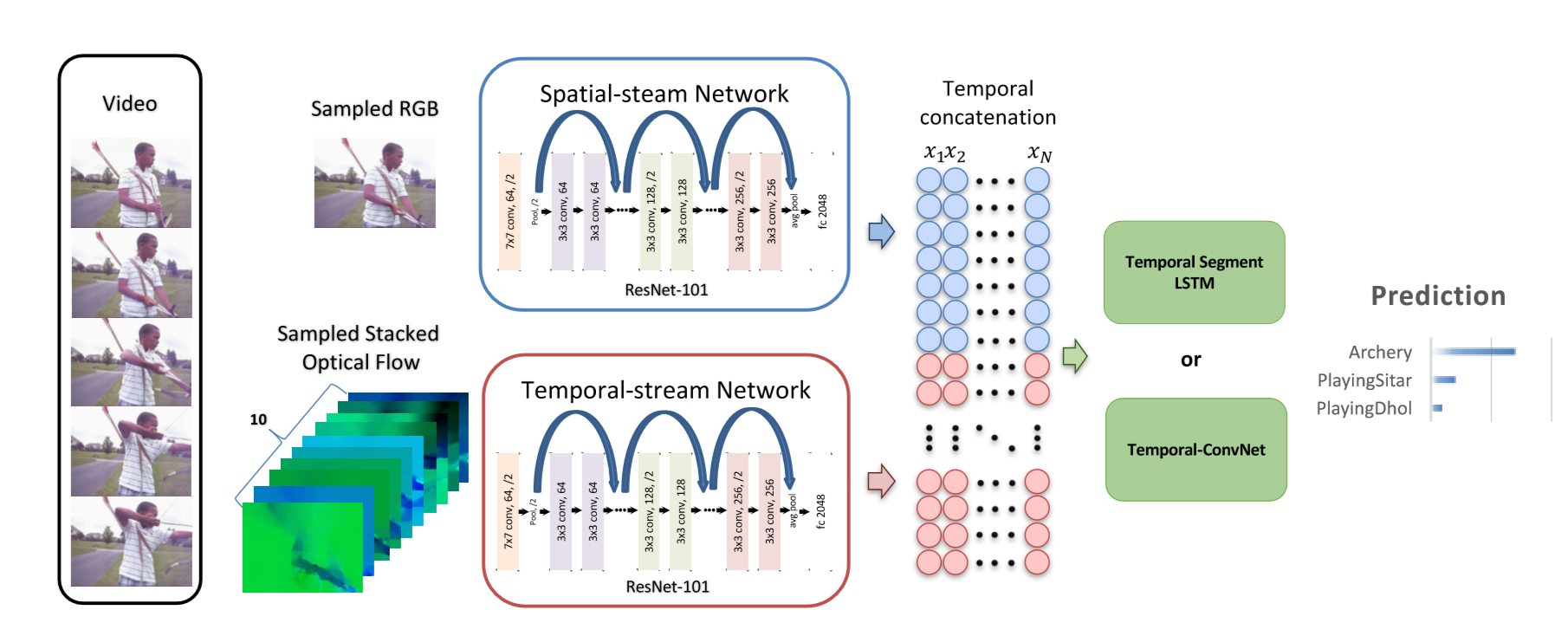}
\caption{proposed framework in \cite{RN211} that makes use of motion-based features, LSTMs, and Temporal-ConvNets to exploit spatiotemporal dynamics.}
\label{fig:image8}
\end{figure}

Conv3D + motion + RNN: In \cite{RN211}, Ma et al. make use of both LSTMs and Temporal-ConvNets. In this method, spatial and temporal features are extracted from a two-stream ConvNet using ResNet-101 pre-trained on ImageNet and ﬁne-tuned for single-frame activity prediction. The spatial-stream network takes RGB images as input, while the temporal-stream network takes stacked optical ﬂow images as inputs. Spatial and temporal features are concatenated and temporally constructed into feature matrices. The constructed feature matrices are then used as input to both proposed methods: temporal segment LSTM (TS-LSTM) and Temporal-Inception (see Figure \ref{fig:image8}). In \cite{RN212}, information from a video is integrated into a map called a motion map using a deep 3D convolutional network. A motion map and the next video frame can be integrated into a new motion map. This technique can be trained by increasing the training video length iteratively. The acquired network can be used for generating the motion map of the whole video. Next, a linear weighted fusion scheme is used to fuse the network feature maps into spatio-temporal features. Finally, a long short-term memory encoder-decoder is used for final predictions.

\subsubsection{Depth}
Conv3D + motion: In \cite{RN213}, 3D dynamic voxel (3DV) is proposed to facilitate depth-based 3D action recognition. The key idea of 3DV is to encode 3D motion information within depth video into a regular voxel set (i.e., 3DV), via temporal rank pooling. Each available 3DV voxel intrinsically involves both 3D spatial and motion features. 3DV is then represented as a point set and input into PointNet++ \cite{RN214} for 3D action recognition. In addition, a multi-stream 3D action recognition manner is also proposed to learn motion and appearance features jointly. To extract richer temporal order information of actions, the depth video is divided into temporal splits and encode this procedure in 3DV integrally.

\subsubsection{Skeleton}
Motion + RNN: In \cite{RN215}, a feature selection network (FSN) is proposed with actor-critic reinforcement learning. Given the extracted feature sequence, FSN learns to adaptively select the most representative features and discard the ambiguous features for action recognition. In addition, a generalized graph generation module is proposed to capture latent dependencies and further propose a generalized graph convolution network (GGCN). The GGCN and FSN are combined in a three-stream recognition framework, in which different types of information from skeleton data are fused to improve recognition accuracy. The proposed method in \cite{RN216} consists of three major steps; feature extraction from a skeleton sequence (as the input of ten neural networks; three LSTM models and seven CNN models), neural networks training, and the late score fusion. Applying various methods to a skeleton sequence can obtain various features prominently in the spatial or temporal domain, and that features are defined as SPF (spatial-domain-feature) and TPF (temporal-domain-feature). SPF is selected as the input of LSTM networks and TPF as the input of CNN networks. To obtain the SPF, three types of spatial domain features are extracted including R (relative position), J (distances between joints), and L (distances between joints and lines). To obtain the TPF, the methods used in \cite{RN19} and \cite{RN13} are followed to generate the joint distances map and joint trajectories map (JTM) respectively.
Conv3D + motion: In \cite{RN217}, dynamic GCN is proposed in which a convolutional neural network named context-encoding network (CeN) is introduced to learn skeleton topology. In particular, when learning the dependency between two joints, contextual features from the rest joints are incorporated in a global manner. Multiple CeN-enabled graph convolutional layers are stacked to build dynamic GCN. In addition, static physical body connections and motion modalities are combined to improve results. In \cite{RN218,RN219}, a two-stream model using 3D CNN is proposed for skeleton-based action recognition. In this method, skeleton joints are mapped into a 3D coordinate space and the spatial and temporal information are encoded. Second, 3D CNN models are separately applied to extract deep features from two streams. Third, to enhance the ability of deep features to capture global relationships, every stream is extended into a multitemporal version. In \cite{RN220}, a data reorganizing strategy is proposed to represent the global and local structure information of human skeleton joints. It employs the data mirror to increase the relationship between skeleton joints. Based on this design, an end-to-end multi-dimensional CNN network is proposed to consider the spatial and temporal information to learn the feature extraction transform function. Particularly, in this CNN network, different convolution kernels are used on different dimensions to learn skeleton representation to generate robust features.

\subsubsection{RGB \& Depth}
Conv3D + motion + RNN: In \cite{RN221},  weighted dynamic images, created from the depth and RGB videos are inputs of the framework. The framework is composed of bidirectional rank pooling, CNNs, and 3D ConvLSTM to extract complementary information from weighted dynamic images. Canonical correlation analysis is employed for feature-level fusion, and a linear SVM is used for predicting the class of action.  In \cite{RN222}, a three-stream framework via 3D CNN, ConvLSTM, 2D CNN, temporal pooling, and a fully connected layer with softmax is employed to extract spatio-temporal features of RGB, depth, and optical flow modalities. In \cite{RN223}, Molchanov et al. employed 3D CNNs and RNNs in the proposed framework for hand gesture recognition using RGB, optical flow, depth, IR, and IR disparity modalities. Each modality's class conditional probability vectors are averaged and fused to detect and classify hand gestures.

motion + RNN: Dhiman et al. \cite{RN224} proposed motion and shape temporal dynamics (STD) as action cues. They employed a framework with RGB dynamic images in motion stream and depth silhouette in STD stream for recognizing action from an unknown view.

Conv3D + motion: In \cite{RN225} and \cite{RN226}, a spatio-temporal attention framework is presented to identify the most representative regions and frames in a video. Following this, RGB, flow, and depth features are retrieved by the ResC3D network and are fused using canonical correlation analysis. A linear SVM classiﬁer estimates the class label. Duan et al.\cite{RN227} propose a four-stream network for gesture recognition. This method uses a two-stream convolutional consensus voting network (2SCVN) for the RGB stream and optical ﬂow ﬁelds to model short and long-term video sequences. Besides, a two-stream 3D depth-saliency ConvNet (3DDSN) is employed to learn fine-grain motion and eliminate background clutter. Some approaches try to predict pose from RGB data and use it in action prediction. In \cite{RN228}, a framework is provided for hierarchical region-adaptive multi-time resolution depth motion map (RAMDMM) and multi-time resolution RGB action recognition system. The suggested approach presents a feature representation method for RGB-D data that allows multi-view and multi-temporal action recognition. Original and synthesized viewpoints are employed for multi-view human action recognition. In addition, to be invariant to changes in an action’s speed, it also employs temporal motion information by incorporating it into the depth sequences. Appearance information in terms of multi-temporal RGB data is utilized to emphasize the underlying appearance information that would otherwise be lost with depth data alone, which helps to enhance sensitivity to interactions with tiny objects. Wu et al. applied 3D CNNs with multimodal inputs to improve spatio-temporal features \cite{RN229}. This method suggests two distinct video presentations; depth residual dynamic image sequence (DRDIS) and pose estimation map sequence (PEMS). DRDIS displays spatial motion changes of an action over time which is robust under lighting conditions, texture, and color variations. PEMS is created by pose (skeletal) estimation from an RGB video and removes the background clutter. In \cite{RN230}, a gesture recognition framework called MultiD-CNN is suggested to learn spatio-temporal features from RGB-D videos. This method includes spatial and temporal information using two recognition models: 3D-CDCN and 2D-MRCN. 3D-CDCN adds the temporal dimension and makes use of 3D ResNets and ConvLSTM to concurrently learn spatio-temporal features. On the other hand, 2D-MRCN collects the motion throughout the video sequences into a motion representation and employs 2D ResNets to learn.
Conv3D +  RNN:  In \cite{RN231}, a two-stream network is suggested to extract spatio-temporal features which are more robust to background clutter from RGB-D videos. The framework contains 3D CNN, convolutional LSTM, spatial pyramid pooling, and a fully connected layer to provide better long-term spatio-temporal learning. 

\subsubsection{RGB \& Skeleton}
motion + RNN: Song et al. \cite{RN232} offered an end-to-end trainable three-stream framework from RGB and optical flow videos. The framework employs skeleton data as a guide for the RGB stream besides skeleton data is trained in a separate stream. The framework is created from a ConvNet with LSTM. Visual features around critical joints are extracted automatically using a skeleton-indexed transform layer, and via a part-aggregated pooling. The visual features of different body parts and actors are uniformly regulated. 
Conv3D + motion: In \cite{RN233}, a fusion-based action recognition framework is proposed. The suggested framework is composed of three parts, including 3DCNN, human skeleton manifold representation, and classifier fusion.  In \cite{RN234}, a multi-stream attention-enhanced adaptive graph convolutional neural network is proposed for skeleton-based action recognition. The graph topology of the skeleton data is learned adaptively in this model. Besides, a spatial-temporal-channel (STC) attention module is embedded in every graph convolutional layer, which helps focus on more important joints, frames, and features. The joints, bones, and the corresponding motion information are modeled in a unified multi-stream framework. In addition, the skeleton data is fused with the skeleton-guided cropped RGB data, which brings additional improvement.

Conv3D + attention: Das et al. \cite{RN235,RN236} propose video-pose networks (VPN and VPN++) for the recognition of activities of daily living. VPN requires both RGB and 3D poses to classify actions. The RGB images are processed by a visual backbone that generates a spatio-temporal feature map. The VPN that takes as input the feature map and the 3D poses consists of two components: an attention network and a spatial embedding. The attention network further consists of a Pose Backbone and a spatio-temporal Coupler. VPN computes a modulated feature map that is used for classification. VPN++ is an extension of VPN to transfer the pose knowledge into RGB through a feature-level distillation and to mimic pose-driven attention through an attention-level distillation. Features of inputs are extracted via two distinct videos and pose backbones. The video backbone consists of 3D CNNs to extract spatio-temporal features, and the pose backbone contains a spatio-temporal GCN. 

\begin{figure}
\centering
\includegraphics[width=0.98\columnwidth]{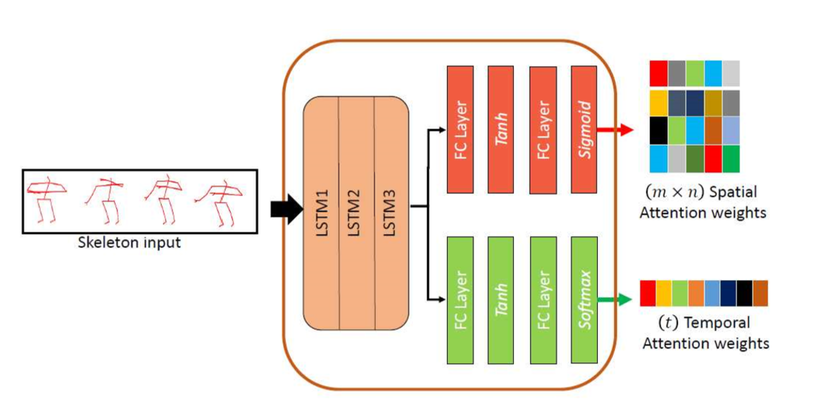}
\caption{A detailed picture of pose-driven RNN attention model in \cite{RN237}}
\label{fig:image9}
\end{figure}

Conv3D +  RNN + attention: In \cite{RN237}, an end-to-end separable spatio-temporal attention network is proposed. The input of the network is human body tracks of RGB videos and their 3D poses. Two separate branches are dedicated to spatial and temporal attention individually. Finally, both branches are combined to classify the activities. Figure \ref{fig:image9} shows a detailed picture of pose driven RNN attention model which takes 3D pose input and computes m × n spatial and t temporal attention weights for the t × m × n × c spatio-temporal features from I3D. In this work, skeleton data is employed as a guide for the RGB stream. Besides, skeleton data participates in the learning itself in a stream.

\subsubsection{Discussions }
Different arrangements are probable for the hybrid techniques (11 combinations). However, some approaches are more established; such as the combination of motion-based features along with 3D filters or LSTMs and Conv3D. In brief, using Conv3D is common in hybrid methods while the combination with the transformers is a new approach. Applying various temporal modeling methods to a sequence can obtain the advantages of different approaches. For example, for motion-based + Conv3D, using motion-based features is a simple approach to obtain a global sense of motion and transactions among consecutive frames. These transactions may be learned through time using 3D filters. In this way, the motion is learned hierarchically through different approaches. However, there are limited methods for these hybrid approaches, and how these approaches should be combined is not well explored.
%%%%%%%%%%%%%%%%%%%%%%%%%%%%%%%%%%%%%%%

\section{Discussions and Future Prospects }
\label{Discussions and Future Prospects}
In brief, methods can learn temporal features by 3D ﬁlters in their 3D convolutional and pooling layers. 3D ConvNets are straightforward extensions of 2D ConvNets as they capture temporal information using 3D convolutions. One limitation of 3D ConvNets is that they typically consider very short temporal intervals, thereby failing to capture long-term temporal information. It has been also shown that using training networks on pre-computed motion features is a way to implicitly learn motion features. These networks generally utilize multiple convolutional networks to model both appearance and motion information in action videos. In this way, pre-trained 2D ConvNets can be utilized, allowing networks that are ﬁne-tuned on stacked optical ﬂow frames to achieve good performance despite limited training data. However, the major problem in the multi-stream networks is that they do not allow interactions among the streams while such an interaction is important for learning spatiotemporal features. Temporal models like RNN and LSTM cope with longer-range temporal relations but the problem with RNNs, and LSTMs, is that it’s hard to parallelize the work for processing sequences. The essential advantage of approaches in the transformer group is that they do not suffer from long dependency issues. The transformers process a sequence as a whole in parallel, instead of using past hidden states to capture dependencies in RNN and LSTM. This training parallelization allows training on larger datasets than was once possible. In addition, there is no risk to lose (or "forget") past information. Moreover, multi-head attention and positional embeddings both provide information about the relationship between different elements. Note that although transformers have the potential of learning longer-term dependency, but are limited by a fixed-length context. Some studies such as Transformer-XL architecture \cite{RN238} try to learn dependency beyond a fixed length without disrupting temporal coherence. Finally, fusing various temporal modeling methods can obtain the advantages of different approaches. However, there are limited methods for these combinations, and how these approaches should combine is not well explored. Table \ref{tab:pros and cons} lists the pros and cons of deep-based temporal modeling approaches.

\begin{table}
\scriptsize
\centering
\caption{pros and cons of deep-based temporal modeling approaches.}
\label{tab:pros and cons}
\begin{tabular}{|l|p{2.5cm}|p{2.5cm}|p{2.5cm}|p{2.5cm}|p{2.5cm}|}
\toprule
  & Motion-based & 3D-CNN & RNN/LSTM & Transformer & Hybrid 
\\ \midrule\midrule
Pros & 
-Pre-trained 2D-ConvNets can be utilized
 & -A natural extension of 2D ConvNets.

-Quite fast to train  and effective with short sequences.

-Can capture inductive biases such as translation equivariance and locality
&
-Modeling longer-range temporal relations

-Past information is retained through past hidden states.
&
-Do not suffer from long dependency issues

-Requiring signiﬁcantly less computational and memory resources

-Requiring signiﬁcantly less computational and memory resources

-High speed in training and inference
 &
-Benefiting from advantages of different approaches

-Hierarchical learning of features

-Parallel processing 

\\ \midrule
Cons & -The High computational cost of computing accurate optical flow 

-Interaction among different streams is difficult. & - Short temporal  interval  

-demanding many computational resources in the training stage 

-Rigid in capturing action sequences with fine-grained visual patterns & -Sequential processing 

-Past information retained through past hidden states 

-Vanishing gradients problems 

-High number of learnable parameters & -Training transformer-based architectures can be expensive, especially for long sequences. 

-Requiring large-scale training to surpass inductive bias & -Limited studies

\\ \bottomrule 	
\end{tabular}
\end{table}

\subsection{Which approaches and modalities are more common? }
\begin{figure}
\centering
\includegraphics[width=0.98\columnwidth]{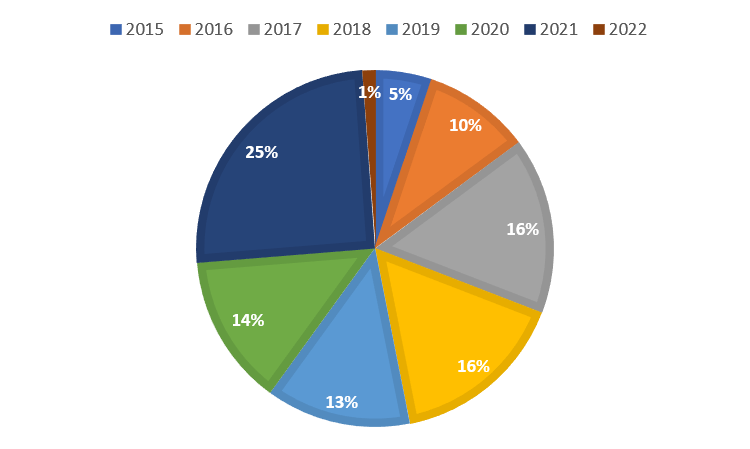}
\caption{the number of studied papers in each year from 2015 to 2022.}
\label{fig:image10}
\end{figure}

In total, more than 170 papers on human action recognition are reviewed in this study from 2015 to 2022 (see Table \ref{tab:papers}). In addition, Figure \ref{fig:image10} shows the number of studied papers in each year. 

\begin{table}
\scriptsize
\centering
\caption{All papers reviewed in this study.}
\label{tab:papers}
\begin{tabular}{|c|c|c|}
\toprule
Time Modeling Approach & Modality & Paper
\\ \hline
\multirow{7}{*}{Motion-based}& RGB& \cite{RN219,RN73,RN74,RN75,RN76,RN77,RN239}\cite{RN78,RN79,RN80, RN81}\\\
& Depth & \cite{RN78,RN79,RN80,RN81} \\\
& Skeleton & \cite{RN82,RN83,RN84,RN85,RN86,RN87,RN88} \\\
& RGB+Depth & \cite{RN89,RN90,RN93,RN94,RN95,RN96,RN97,RN99,RN100,RN101,RN102,RN103,RN104,RN105,RN106,RN107} \\\
& RGB+Skeleton & \cite{RN108,RN109} \\\
& Depth+Skeleton & \cite{RN110,RN111,RN112}  \\\
& RGB+Depth+Skeleton & \cite{RN113,RN114,RN115,RN116,RN117,RN118} \\\hline
\multirow{7}{*}{Three Dimensional Filters} & RGB & \cite{RN17,RN18,RN119,RN120,RN121,RN122,RN123,RN240,RN241,RN242,RN243} \\\
& Depth & \cite{RN124} \\\
& Skeleton & \cite{RN125,RN126,RN127,RN244,RN245,RN246} \\\
& RGB+Depth & \cite{RN128,RN129,RN130,RN131,RN132,RN133} \\\
& RGB+Skeleton & \cite{RN247,RN248} \\\
& Depth+Skeleton & \cite{RN134}  \\\
& RGB+Depth+Skeleton & \\\hline
\multirow{7}{*}{Temporal Sequence Models} & RGB & \cite{RN20,RN139,RN140} \\\
& Depth & \cite{RN141,RN142,RN143} \\\
& Skeleton & \cite{RN144,RN145,RN146,RN147,RN148,RN149,RN150,RN151,RN152,RN153,RN154,RN249} \\\
& RGB+Depth & \cite{RN155,RN156} \\\
& RGB+Skeleton &  \\\
& Depth+Skeleton & \cite{RN157,RN158,RN159} \\\
& RGB+Depth+Skeleton & \cite{RN160} \\\hline
\multirow{7}{*}{Transformers and Attention} & RGB & \cite{RN171,RN172,RN173,RN174,RN175,RN177,RN178,RN179,RN180,RN182,RN184,RN185,RN187,RN188,RN190,RN250,RN251,RN252,RN253,RN254,RN255,RN256,RN257,RN169,RN191,RN192,RN193,RN258} \\\
& Depth &  \\\
& Skeleton & \cite{RN194,RN195,RN196,RN197,RN198,RN199,RN200,RN201,RN202,RN203,RN204,RN205,RN259} \\\
& RGB+Depth & \cite{RN210} \\\
& RGB+Skeleton & \cite{RN206,RN207,RN208,RN209} \\\
& Depth+Skeleton &   \\\
& RGB+Depth+Skeleton & \\\hline
\multirow{7}{*}{Hybrid Methods} & RGB & \cite{RN211,RN212,RN260,RN261}\\\
& Depth & \cite{RN213} \\\
& Skeleton & \cite{RN215,RN216,RN217,RN218,RN219,RN220,RN262} \\\
& RGB+Depth & \cite{RN221,RN222,RN223,RN224,RN225,RN226,RN227,RN228,RN229,RN230,RN231} \\\
& RGB+Skeleton & \cite{RN232,RN233,RN234,RN235,RN236,RN237} \\\
& Depth+Skeleton &  \\\
& RGB+Depth+Skeleton &
\\ \bottomrule 	
\end{tabular}
\end{table}

\begin{figure}
\centering
\includegraphics[width=0.98\columnwidth]{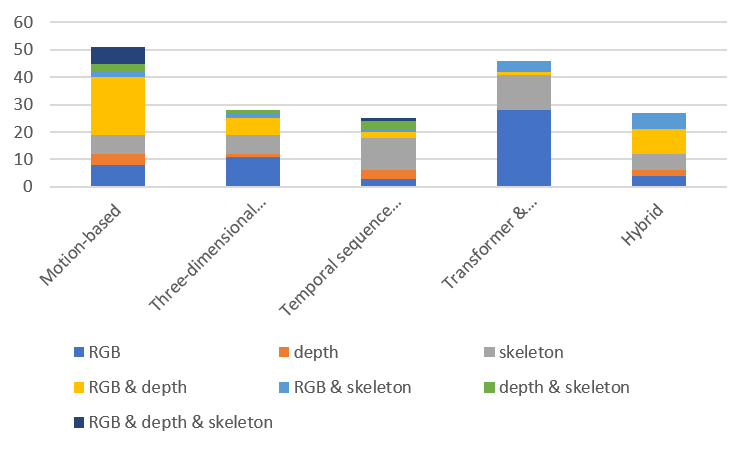}
\caption{The number of studied papers in each category.}
\label{fig:image11}
\end{figure}

\begin{figure}
\centering
\includegraphics[width=0.98\columnwidth]{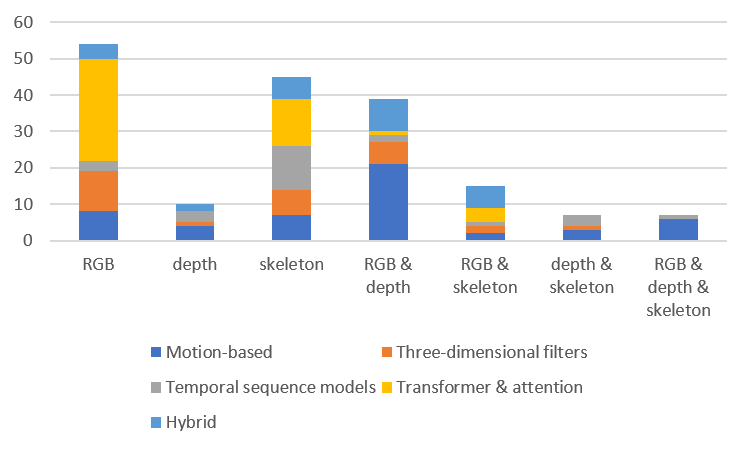}
\caption{The number of studied papers in each modality and their combinations.}
\label{fig:image12}
\end{figure}

 Figure \ref{fig:image11} shows the number of studied papers in each category. As this figure shows, the greatest number of methods use motion-based features for temporal modeling.  In this category, RGB \& depth is the popular used modality. For the categories of Conv3D and transformer, RGB is the common modality used among studied papers while temporal sequence models are used mostly along with the skeleton. This figure is represented from another view in Figure \ref{fig:image12} where the numbers of studied papers in each modality and their combinations are shown. As this figure shows, RGB is the most popular modality among studied papers, since it contains rich information about the appearance and is easy to collect. However, methods based on the RGB modality are often sensitive to viewpoint variations and background clutters, etc. Hence, action recognition with other modalities, such as 3D skeleton data, has also received great attention. As Figure \ref{fig:image12} shows, the skeleton is in the second rank. The skeleton provides the body structure information (as a simple, efficient, and informative representation of human behaviors). Nevertheless, action recognition using only skeleton data still faces challenges, due to its sparse representation, the noisy skeleton information, and the lack of shape information, especially for handling human-object interactions. So, the combination of multiple modalities is used frequently in the literature. In Figure \ref{fig:image12}, The combination of RGB \& depth is in the third rank. It is the most popular combination among all multi-modal approaches. The RGB and depth videos respectively capture the rich appearance and 3D shape information, that are complementary and can be used for action recognition. 

\begin{figure}
\centering
\includegraphics[width=0.98\columnwidth]{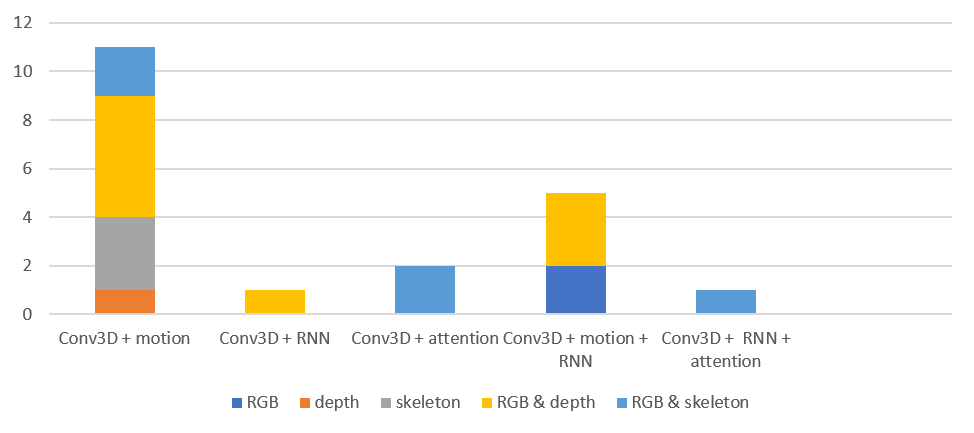}
\caption{the number of studied papers for hybrid methods.}
\label{fig:image13}
\end{figure}

For hybrid methods, the try is to learn temporal features by combining different approaches. While different combinations are possible, some are more promising. Figure \ref{fig:image13} shows the number of studied papers in each combination. Note that only five combinations are shown in this figure because the remaining combinations did not include any paper among the studied papers. Note that, Conv3D is present in all five combinations since the 3D CNN-based methods are very powerful in modeling discriminative features from both the spatial and temporal dimensions. In addition, as this figure shows, Conv3D + motion is used a lot in the literature. Since utilizing motion-based features is also a common approach for modeling temporal variations, the combination of these two approaches (Conv3D and motion) may improve accuracy. 

\subsection{Which approaches and modalities are the winner? }
To compare different approaches with each other, six benchmark datasets of human action recognition including NTU RGB+D \cite{RN149}, NTU RGB+D 120 \cite{RN263}, Toyota-Smarthome \cite{RN237}, Kinetics 400 \cite{RN264}, Kinetics 600 \cite{RN265}, and Skeleton-Kinetics are selected. As one of the most fundamental tasks in computer vision, there are numerous benchmark datasets for unimodal or multimodal vision-based human action recognition \cite{RN266,RN267,RN268,RN269}. These benchmark datasets are chosen due to their popularity and large number of actions. While the three first datasets are multimodal, the Kinetics 400 and 600 provide only RGB data and Skeleton-Kinetics include only skeletal data. 
NTU RGB+D \cite{RN149}: is a large-scale multimodal human action recognition dataset containing 56,880 action sequences of 60 action classes. The action samples are performed by 40 persons in the lab environment and are captured by three Microsoft Kinect v2 cameras from three different views. Each sample contains an action and contains at most 2 subjects. We report the two standard evaluation protocols recommended by the authors of this dataset namely cross-subject (CS) and cross-view (CV). In the CS setting, training data comes from 20 subjects and test data comes from the other 20 subjects. In the CV setting, training data comes from camera views 2 and 3, and test data comes from camera view 1. This dataset contains RGB videos, depth map sequences, 3D skeletal data, and infrared (IR) videos for each sample.
NTU RGB+D 120 \cite{RN263}: extends NTU RGB+D with additional 57,600 sequences over 60 extra action classes. Totally 114,480 samples over 120 classes are performed by 106 individuals, captured with three camera views. There are two recommended evaluation protocols, namely cross-subject (CS) and cross setup (CSet). In the CS setting, 63,026 clips from 53 subjects are used for training, and the remaining subjects are reserved for testing. In the CSet setting, 54,471 clips with even setup IDs are used for training, and the rest clips with odd setup IDs are used for testing.
Toyota-Smarthome \cite{RN237}: is a dataset of activities of daily living recorded in an apartment where 18 older subjects carry out tasks of daily living during a day. The dataset contains 16.1k video clips, 7 different camera views, and 31 complex activities performed in a natural way without strong prior instructions. This dataset provides RGB data and 3D skeletons which are extracted from LCRNet \cite{RN270}. For the evaluation of this dataset, the cross-subject (CS) and two cross-view protocols (CV1 and CV2) \cite{RN237} are reported. 
Kinetics \cite{RN264,RN265}: is a collection of large-scale, high-quality datasets of URL links consisting of 10-second videos sampled at 25fps from YouTube. Here both Kinetics 400 \cite{RN264} and 600 \cite{RN265} are considered, containing respectively 400 and 600 classes. Note that there is a relatively newer version of this dataset called Kinetics-700 \cite{RN271} that is not considered here, because existing results are limited on this new version. As these are dynamic datasets and videos may be removed from YouTube, the size of these datasets are approximately 267k and 446k respectively. Kinetics-400 consists of $\sim$240k training videos and 20k validation videos. Kinetics-600 has $\sim$392k training videos and 30k validation videos. In addition, Skeleton-Kinetics is derived from the Kinetics-400 dataset. The skeletons are estimated by \cite{RN272} from RGB videos using the OpenPose toolbox \cite{RN273}. Each joint consists of 2D coordinates in the pixel coordinate system and its confidence score. There are 18 joints for each person. In each frame at most two subjects are considered. Skeleton-based approaches reported on Kinetics usually use Skeleton-Kinetics for evaluation. Here the same train-validation split as \cite{RN272} is considered. That is, the training and validation sets contain 240k and 20k video clips respectively. Top-1 accuracies are reported.
Table \ref{tab:STOA} shows state-of-the-art approaches on these benchmark datasets. As this table shows, Conv3D or Conv3D and its combination with attention is the top time modeling approach for NTU RGB+D, NTU RGB+D 120, and Toyota-Smarthome, all using RGB \& skeleton as the input modalities. For Kinetics-400 and Kinetics-600 attention is the winner. However, for Kinetics-Skeleton, Conv3D is still at the first rank. Note that multimodal methods achieve superior results compared with unimodal methods on all three first multimodal datasets. Finally, Conv3D and attention are the most popular approaches among state-of-the-art methods for modeling temporal variations in human action recognition. Note that for RGB-based methods using pure-transformer architectures is becoming a common approach showing the capability and the increasing interest of the community in using transformers for temporal modeling. In addition, according to the table, the combination of RGB and pose is more frequent among top multi-modal human action recognition algorithms as they provide complementary information about the appearance and 3D coordinates of joints.

\scriptsize
\begin{center}
\begin{longtable}[]{lclclclclc|}
\caption {Top state-of-the-art methods on six benchmark datasets.} 
\label{tab:STOA} \\
\hline
\toprule
\begin{minipage}[b]{0.15\columnwidth}\raggedright\strut
\textbf{Dataset}\strut
\end{minipage} & \begin{minipage}[b]{0.2\columnwidth}\raggedright\strut
\textbf{Method}\strut
\end{minipage} & \begin{minipage}[b]{0.21\columnwidth}\raggedright\strut
\textbf{Accuracy}\strut
\end{minipage} & \begin{minipage}[b]{0.15\columnwidth}\raggedright\strut
\textbf{Modality}\strut
\end{minipage} & \begin{minipage}[b]{0.21\columnwidth}\raggedright\strut
\textbf{Time modeling}\strut
\end{minipage}\tabularnewline
\midrule
\endhead
\begin{minipage}[t]{0.15\columnwidth}\raggedright\strut
NTU RGB+D \cite{RN149}\strut
\end{minipage} & \begin{minipage}[t]{0.2\columnwidth}\raggedright\strut
Duan et al. \cite{RN247}\newline Das et al. \cite{RN236}\newline Shi et al. \cite{RN234}\newline
Davoodikakhki et al. \cite{RN248} \newline Das et al. \cite{RN235}\newline  Zhu et al. \cite{RN240} \newline
Das et al. \cite{RN208}\newline  Chen et al. \cite{RN125} \newline Das et al. \cite{RN237}\newline
Piergiovanni et al. \cite{RN241}\newline Liu et al. \cite{RN126}\newline Ye et al.
\cite{RN217} \strut
\end{minipage} & \begin{minipage}[t]{0.21\columnwidth}\raggedright\strut
97.0 (cs) 99.6 (cv) \newline 96.6 (cs) 99.1 (cv) \newline 96.1 (cs) 99.0 (cv) \newline
95.66(cs)98.79(cv) \newline 95.5 (cs) 98.0 (cv) \newline 94.3 (cs) 97.2 (cv) \newline
93.0 (cs) 95.4 (cv) \newline 92.4 (cs) 96.8 (cv) \newline 92.2 (cs) 94.6 (cv) \newline 
              93.7 (cv) \newline 91.5 (cs) 96.2 (cv) \newline 91.5 (cs) 96.0 (cv)\strut
\end{minipage} & \begin{minipage}[t]{0.15\columnwidth}\raggedright\strut
RGB \& skeleton \newline RGB \& skeleton \newline RGB \& skeleton \newline RGB \& skeleton \newline 
RGB \& skeleton \newline RGB \newline RGB \& skeleton \newline skeleton \newline RGB \& skeleton \newline RGB \newline skeleton \newline
skeleton\strut
\end{minipage} & \begin{minipage}[t]{0.21\columnwidth}\raggedright\strut
Conv3D \newline Conv3D+attention \newline Conv3D+motion \newline Conv3D \newline Conv3D+attention \newline
Conv3D \newline attention \newline Conv3D \newline Conv3D+RNN+attention \newline Conv3D \newline Conv3D \newline Conv3D+motion\strut
\end{minipage}\tabularnewline
\hline
\begin{minipage}[t]{0.15\columnwidth}\raggedright\strut
NTU RGB+D 120 \cite{RN263}\strut
\end{minipage} & \begin{minipage}[t]{0.2\columnwidth}\raggedright\strut
Duan et al. \cite{RN247} \newline Das et al. \cite{RN236} \newline Chen et al. \cite{RN125} \newline Ye et
al. \cite{RN217} \newline Das et al. \cite{RN235} \newline Das et al. \cite{RN237} \newline Papadopoulos et
al. \cite{RN244} \newline Caetano et al. \cite{RN85} \newline Caetano et al. \cite{RN84} \newline Liu et al.
\cite{RN274}\strut
\end{minipage} & \begin{minipage}[t]{0.21\columnwidth}\raggedright\strut
95.3(cs) 96.4(cset) \newline 90.7(cs) 92.5(cset) \newline 88.9(cs) 90.6(cset) \newline 
87.3(cs) 88.6(cset) \newline 86.3(cs) 87.8(cset) \newline 83.8(cs) 82.5(cset) \newline 
78.3(cs) 79.2(cset) \newline 67.9(cs) 62.8(cset) \newline 66.9(cs) 67.7(cset) \newline 
64.6(cs) 66.9(cset)\strut
\end{minipage} & \begin{minipage}[t]{0.15\columnwidth}\raggedright\strut
RGB \& skeleton \newline RGB \& skeleton \newline skeleton \newline skeleton \newline RGB \& skeleton \newline RGB \&
skeleton \newline skeleton \newline skeleton \newline skeleton \newline skeleton\strut
\end{minipage} & \begin{minipage}[t]{0.21\columnwidth}\raggedright\strut
Conv3D \newline Conv3D+attention \newline Conv3D \newline Conv3D+motion \newline Conv3D+attention \newline
Conv3D+RNN+attention \newline Conv3D \newline motion \newline motion \newline Conv2D\strut
\end{minipage}\tabularnewline
\hline
\begin{minipage}[t]{0.15\columnwidth}\raggedright\strut
Toyota-Smarthome \cite{RN237}\strut
\end{minipage} & \begin{minipage}[t]{0.2\columnwidth}\raggedright\strut
Das et al. \cite{RN236} \newline Yang et al. \cite{RN259} \newline Ryoo et al. \cite{RN260} \newline
Kangaspunta et al. \cite{RN261} \newline Das et al. \cite{RN235} \newline Das et al. \cite{RN237} \newline
Wang et al. \cite{RN258} \newline Carreira et al. \cite{RN121} \newline Mahasseni et al.
\cite{RN249} \newline Ohnishi et al. \cite{RN239}\strut

\end{minipage} & \begin{minipage}[t]{0.21\columnwidth}\raggedright\strut
71.0(cs) 58.1(cv2) \newline 64.3(cs) 36.1(cv1) 65.0(cv2) \newline 63.6(cs) \newline 62.11(cs) \newline
60.8(cs) 53.5(cv2) \newline 54.2(cs) 35.2(cv1) 50.3(cv2) \newline 53.6(cs) 34.3(cv1)
43.9(cv2) \newline 53.4(cs) 34.9(cv1) 45.1(cv2) \newline 42.5(cs) 13.4(cv1) 17.2(cv2) \newline
41.9(cs) 20.9(cv1) 23.7(cv2)\strut

\end{minipage} & \begin{minipage}[t]{0.15\columnwidth}\raggedright\strut
RGB \& skeleton \newline skeleton \newline RGB \newline RGB \newline RGB \& skeleton\newline RGB \& skeleton \newline RGB \newline RGB \newline
skeleton \newline RGB\strut
\end{minipage} & \begin{minipage}[t]{0.21\columnwidth}\raggedright\strut
Conv3D+attention \newline attention \newline motion+attention \newline Conv3D+motion \newline Conv3D+attention \newline Conv3D+RNN+attention \newline attention \newline Conv3D \newline RNN \newline motion\strut
\end{minipage}\tabularnewline
\hline
\begin{minipage}[t]{0.15\columnwidth}\raggedright\strut
Kinetics-400 \cite{RN264}\strut
\end{minipage} & \begin{minipage}[t]{0.2\columnwidth}\raggedright\strut
Yan et al. \cite{RN250} \newline Zhang et al. \cite{RN251} \newline Wei et al. \cite{RN252} \newline Yuan et
al. \cite{RN253} \newline Liu et al. \cite{RN254} \newline Li et al. \cite{RN255} \newline Tong et al.
\cite{RN256} \newline Ryoo et al. \cite{RN257} \newline Arnab et al. \cite{RN193} \newline Duan et al.
\cite{RN247} \newline Fan et al. \cite{RN192} \newline Bertasius et al. \cite{RN191} \newline Feichtenhofer
et al. \cite{RN242} \newline Feichtenhofer et al. \cite{RN243}\strut
\end{minipage} & \begin{minipage}[t]{0.21\columnwidth}\raggedright\strut
89.1 (top-1) \newline 87.2 (top-1) \newline 87.0 (top-1) \newline 86.8 (top-1) \newline 86.8 (top-1) \newline 86.1
(top-1) \newline 85.8 (top-1) \newline 85.4 (top-1) \newline 84.9 (top-1) \newline 83.9 (top-1) \newline 81.2 (top-1)
\newline 80.7 (top-1) \newline 80.4 (top-1) \newline 79.8 (top-1)\strut
\end{minipage} & \begin{minipage}[t]{0.15\columnwidth}\raggedright\strut
RGB \newline RGB \newline RGB \newline RGB \newline RGB \newline RGB \newline RGB \newline RGB \newline RGB \newline RGB \& skeleton \newline RGB \newline RGB \newline RGB \newline RGB\strut
\end{minipage} & \begin{minipage}[t]{0.21\columnwidth}\raggedright\strut
attention \newline attention \newline attention \newline attention \newline attention \newline attention \newline attention \newline
attention \newline attention \newline Conv3D \newline attention \newline attention \newline Conv3D \newline Conv3D\strut
\end{minipage}\tabularnewline
\hline
\begin{minipage}[t]{0.15\columnwidth}\raggedright\strut
Kinetics-600 \cite{RN265}\strut
\end{minipage} & \begin{minipage}[t]{0.2\columnwidth}\raggedright\strut
Yan et al. \cite{RN250} \newline Wei et al. \cite{RN252} \newline Yuan et al. \cite{RN253} \newline Li et
al. \cite{RN255} \newline Zhang et al. \cite{RN251} \newline Ryoo et al. \cite{RN257} \newline Liu et al.
\cite{RN188} \newline Arnab et al. \cite{RN193} \newline Fan et al. \cite{RN192} \newline Bertasius et al.\cite{RN191}
\newline Feichtenhofer et al. \cite{RN242} \newline Feichtenhofer et al.
\cite{RN243}\strut
\end{minipage} & \begin{minipage}[t]{0.21\columnwidth}\raggedright\strut
89.6 (top-1) \newline 88.3 (top-1) \newline 88.0 (top-1) \newline 87.9 (top-1) \newline 87.9 (top-1) \newline 86.3
(top-1) \newline 86.1 (top-1) \newline 85.8 (top-1) \newline 84.1 (top-1) \newline 82.2 (top-1) \newline 81.9 (top-1)
\newline 81.8 (top-1)\strut
\end{minipage} & \begin{minipage}[t]{0.15\columnwidth}\raggedright\strut
RGB \newline RGB \newline RGB \newline RGB \newline RGB \newline RGB \newline RGB \newline RGB \newline RGB \newline RGB \newline RGB \newline RGB\strut
\end{minipage} & \begin{minipage}[t]{0.21\columnwidth}\raggedright\strut
attention \newline attention \newline attention \newline attention \newline attention \newline attention \newline attention \newline
attention \newline attention \newline attention \newline Conv3D \newline Conv3D\strut
\end{minipage}\tabularnewline
\hline
\begin{minipage}[t]{0.15\columnwidth}\raggedright\strut
Kinetics-Skeleton \cite{RN272}\strut
\end{minipage} & \begin{minipage}[t]{0.2\columnwidth}\raggedright\strut
Duan et al. \cite{RN247} \newline Obinata et al. \cite{RN245} \newline Chen et al. \cite{RN246} \newline Liu
et al. \cite{RN126} \newline Ye et al. \cite{RN217} \newline Shi et al. \cite{RN234} \newline Yang et al.
\cite{RN262} \newline Plizzari et al. \cite{RN201}\strut
\end{minipage} & \begin{minipage}[t]{0.21\columnwidth}\raggedright\strut
47.7 (top-1) \newline 38.6 (top-1) \newline 38.4 (top-1) \newline 38.0 (top-1) \newline 37.9 (top-1) \newline 37.8
(top-1) \newline 37.5 (top-1) \newline 37.4 (top-1)\strut
\end{minipage} & \begin{minipage}[t]{0.15\columnwidth}\raggedright\strut
skeleton \newline skeleton \newline skeleton \newline skeleton \newline skeleton \newline RGB \& skeleton \newline skeleton
\newline skeleton\strut
\end{minipage} & \begin{minipage}[t]{0.21\columnwidth}\raggedright\strut
Conv3D \newline Conv3D \newline Conv3D \newline Conv3D \newline Conv3D+motion \newline Conv3D+motion \newline Conv3D+motion \newline attention\strut
\end{minipage}\tabularnewline
\bottomrule
\end{longtable}
\end{center}
\normalsize

\subsection{Future directions and challenges}

We envision transformers will proceed in human action recognition. However, there are main challenges for using transformers in activity recognition that should be addressed by the community. 

\textbf{Transformers with high performance and low resource cost for action recognition:} Compared with CNN models, transformers are usually huge and computationally expensive, and efficient transformers are needed especially for devices with limited resources. So compressing and accelerating transformer models for efficient implementation; specifically, transformers with high performance and low resource cost is an open problem \cite{RN28}. Some works attempt to compress pre-defined transformer models into smaller ones, some others attempt to design compact models directly. The research carried out for efficient implementation includes pruning networks and decomposition \cite{RN284,RN285}, knowledge distillation \cite{RN286}, network quantization \cite{RN287}, and compact architecture design \cite{RN288}. However, models originally designed for NLP, may not be suitable for action recognition. 

\textbf{Transformer models capable of handling a large number of spatiotemporal tokens and extracting conjoint inter and intra-modal features:} In the case of multimodality, 
transformers are used for intramodality spatial and temporal modeling and cross-modality feature fusion. Handling a large number of spatiotemporal tokens extracted from multiple modalities is a concern. Some methods develop several scalable model variants which factorize self-attention across the space, time, and modality dimensions. For example in \cite{RN202}, a spatial self-attention module is used to understand intra-frame interactions between different body parts, and a temporal self-attention module to model inter-frame correlations. The two are combined in a two-stream network. To further explore the rich inter-modal interactions and their effects, cross-modal attention mechanisms that can be seamlessly integrated into the transformer building block are also needed to effectively exploit the complementary nature of all modalities.

\textbf{Transformer models capable of processing multiple tasks of human action recognition in a single model:} In addition, following the success of some new trends in NLP \cite{RN275} and CV \cite{RN276,RN277,RN278} to develop transformer models capable of processing multiple tasks in a single model, we believe that domains including images, audio, multimodal, etc. can be unified in only one model. Advances in hybrid models combining different approaches with transformers are also expected \cite{RN235,RN236}. 

\textbf{Practical action recognition:} Existing approaches, such as 3D convolutional neural networks and transformer-based methods, usually process the videos in a clip-wise manner; requiring huge GPU memory and fixed-length video clips \cite{RN279}. So, proposing models capable of working on variant-length video clips without requiring large GPU memory is needed. 
Finally, we think the deep learning solutions for large-scale, real-time, multi-view, and realistic action recognition topics along with newer problems like early recognition, multi-task learning, few-shot learning, unsupervised and semi-supervised learning, and recognition from low-resolution videos will receive attention in the next years.

%%%%%%%%%%%%%%%%%%%%%%%%%%%%%%%%%%%%%%%
\section{Conclusions}
\label{Conclusions}
In this paper, a comprehensive overview of a long concern in human action recognition i.e. temporal modeling is presented. It is especially important in recognizing similar actions with subtle time differences. The taxonomy is defined to cover most of the basic and crucial approaches for modeling temporal information and then some recent methods are reviewed. Key branches introduced include motion-based feature approaches, three-dimensional convolutional neural networks, recurrent neural networks, transformers, and hybrid methods. In brief, more than 170 recent papers on human action recognition are reviewed. In each category, methods are grouped based on used modalities; RGB, depth, skeleton, or combinations of multiple modalities. Finally, different approaches are compared with each other using benchmark datasets for human action recognition. This way, popular approaches for temporal modeling along with popular modalities are recognized. From studied papers, transformers are showing promising results due to properties such as not suffering from long dependency issues and parallel processing. However, transformers are still at the start point of the way, and the question that transformers will prevail in human action recognition or not remains (due to difficulties such as the expensive cost of training transformer-based architectures, the lack of inductive biases of CNNs in transformers, and the need for large-scale training to surpass these biases).
%%%%%%%%%%%%%%%%%%%%%%%%%%%%%%%%%%%%%%%
\section*{Acknowledgment}
This research is partially supported by the Iran national science foundation (INSF) \href{https://insf.org/en}{https://insf.org/en} and the Shahid Bahonar University of Kerman   \href{https://uk.ac.ir/en/home}{https://uk.ac.ir/en/home} under grant number 98006291.

%%%%%%%%%%%%%%%%%%%%%%%%%%%%%%%%%%%%%
%% The Appendices part is started with the command \appendix;
%% appendix sections are then done as normal sections
\appendix

\section{From self-attention to transformers; formulation overview}
As mentioned before, the transformer that is primarily used in NLP \cite{RN25}, is a new deep learning model based on the self-attention mechanism that weights the significance of different parts of the input data. The self-attention is a key idea behind transformers, which facilities capturing ‘long term’ dependencies between sequence elements and can be viewed as a kind of non-local ﬁltering operation \cite{RN280,RN258}. Note that encoding such dependencies is a challenge in CNNs and RNNs. The main difference between self-attention with convolution is that the ﬁlters are calculated dynamically for any input compared with static ﬁlters of the convolution operation.

\subsection{Building block: self-attention and multi-head attention layer}
A self-attention layer projects the input sequence $X \in R^{n\times d}$
onto three learnable weight matrices namely, Queries, Keys and Values
denoted as $W^{Q} \in R^{d\times d_{q}} $,
$W^{K} \in R^{d\times d_{k}}$ and $W^{V} \in R^{d\times d_{v}}$. Where n is the number of entities in the input
sequence, d is the embedding dimension for representing each entity,
and $d_{q}=d_{k}=d_{v}=d_{\text{model}}$. So the output of the self-attention layer $Z \in R^{n\times d_{v}}$ is obtained as:

\begin{equation}
Q = XW^{Q}, K = XW^{K}, V =XW^{V}
\end{equation}

\begin{equation}
Z=softmax\left(\frac{\text{Q.}K^{T}}{\sqrt{d_{k}}}\right)\text{.V}\ 
\end{equation}

The goal of self-attention is to capture the interaction among all n
entities by computing the scores between each pair of different vectors.
These scores determine the amount of attention given to other entities
when encoding the entity at the current position. Normalizing the scores
enhances gradient stability for improved training, and the softmax
function is used to convert the scores into probabilities. Vectors with
larger probabilities receive additional focus from the following layers.
In this way, each entity is encoded in terms of global contextual
information.

In addition, to improve the performance of the simple self-attention
layer, multi-head attention is used to compute multiple dependencies
among different entities of the sequence. The multi-head attention
contains multiple self-attention blocks, with each block having its own
set of weight matrices $[W^{Q_{i}}, W^{K_{i}},W^{V_{i}}]$, where $i =0\ldots(h-1)$
and $h$ is the number of self-attention blocks. Finally,
the outputs of all blocks are concatenated into a single matrix and
projected onto a weight matrix.

It has been shown in the literature that self-attention (with positional
encodings) is theoretically a more flexible operation \cite{RN281}. In
\cite{RN282}, the relationship between self-attention and convolution
operations is studied. Their empirical results showed that multi-head
self-attention (with sufficient parameters) is a more generic operation
that can model the expressiveness of convolution as a special case.
Self-attention can learn the global as well as local features, and
provide the capability of learning kernel weights adaptively as well as
the receptive field.

\subsection{Transformer model}
\begin{figure}
\centering
\includegraphics[width=0.98\columnwidth]{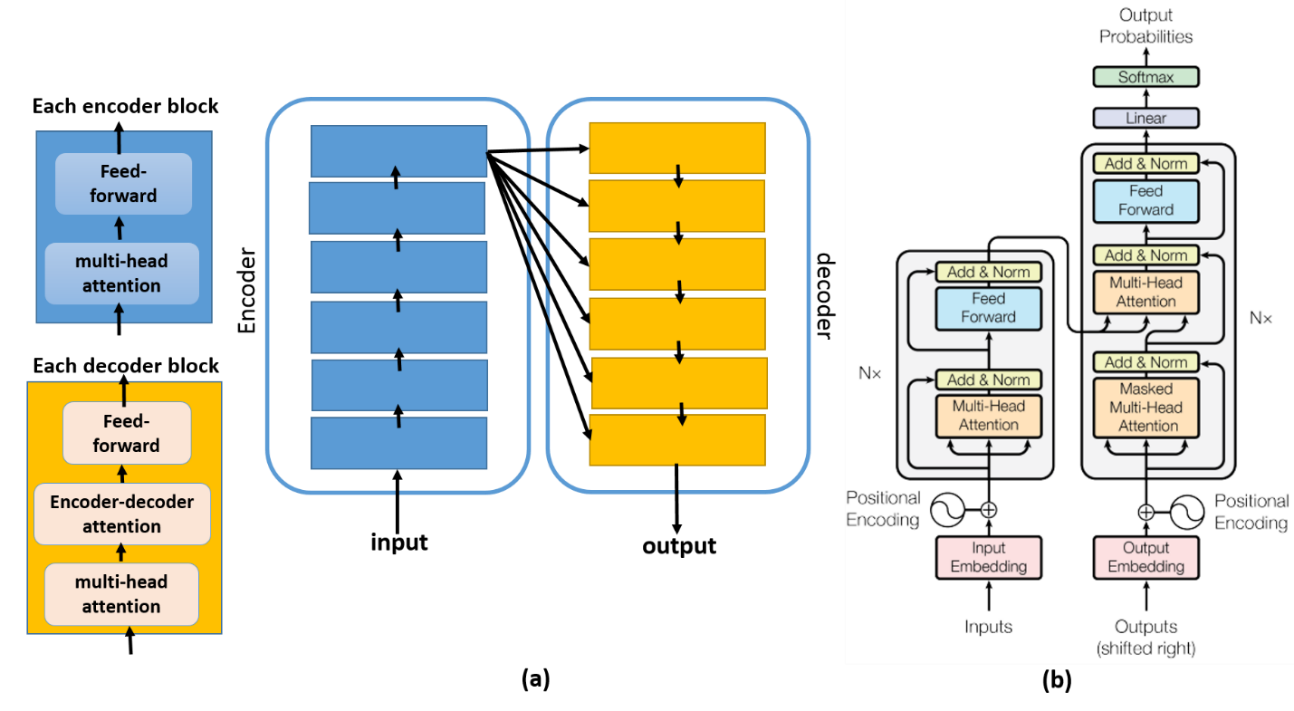}
\caption{The transformer architecture: a) simple schematic and b) more
details of transformer model \cite{RN25}.}
\label{fig:image14}
\end{figure}

The architecture of the transformer model contains an encoder-decoder
structure, as shown in Figure \ref{fig:image14}. The left side of this image shows a
simple schematic of the transformer. The encoder module contains
\(N\)stacked identical blocks, with each block having two
sub-layers of a multi-head self-attention network and a point-wise fully
connected feed-forward network. The decoder in the transformer model
also comprises \(N\) identical blocks. Each decoder block
has three sub-layers of multi-head self-attention, encoder-decoder
attention, and feed-forward. The multi-head self-attention and
feed-forward are similar to the encoder, while the encoder-decoder
attention sublayer performs multi-head attention on the outputs of the
corresponding encoder block.

The right side of Figure \ref{fig:image14} shows more details of the transformer
proposed in \cite{RN25}. Note that After each block in the encoder and
decoder, residual connections \cite{RN186} and layer normalization
\cite{RN283} are also applied. Positional encodings are also added to the
input sequence to capture the relative position of each entity in the
sequence. Since there is no recurrence and convolution in the
transformer model, some information must be embedded about the position
of the entities in the sequence for the model to use the order of the
sequence. Positional encodings have the same dimensions as the input d
and can be learned or pre-defined, e.g., by sine or cosine functions. In
addition, the decoder of the transformer uses previous outputs to
predict the following entity in the sequence. So the decoder takes
inputs from the encoder and the preceding outputs to calculate the next
entity of the sequence.

%% \label{}

%% If you have bibdatabase file and want bibtex to generate the
%% bibitems, please use
%%
\bibliographystyle{elsarticle-num} 
\bibliography{REF}

%% else use the following coding to input the bibitems directly in the
%% TeX file.

%\begin{thebibliography}{00}
%
%%% \bibitem{label}
%%% Text of bibliographic item
%
%\bibitem{}
%
%\end{thebibliography}
\end{document}